\pdfoutput=1

\documentclass[11pt]{article}

\usepackage[final]{acl}

\usepackage{times}
\usepackage{latexsym}
\usepackage{amsmath}
\usepackage[T1]{fontenc}

\usepackage[utf8]{inputenc}

\usepackage{microtype}

\usepackage{inconsolata}
\usepackage[algo2e]{algorithm2e} 
\usepackage{algorithm}
\usepackage{graphicx}

\usepackage{amsmath}

\def\bbE{\mathbb{E}}
\def\bbR{\mathbb{R}}

\def\bx{\textbf{x}}
\def\by{\textbf{y}}

\def\bQ{\textbf{Q}}
\def\bI{\textbf{I}}

\def\\avg{\avg}
\def\cA{\mathcal{A}}

\def\cD{\mathcal{D}}

\def\cT{\mathcal{T}}

\def\cM{\mathcal{M}}

\def\cS{\mathcal{S}}

\def\cX{\mathcal{X}}
\def\cY{\mathcal{Y}}

\def\bx{\boldsymbol{x}}
\def\by{\boldsymbol{y}}

\def\bv{\boldsymbol{v}}
\def\bw{\boldsymbol{w}}

\def\be{\boldsymbol{e}}

\DeclareMathOperator*{\argmin}{arg\,min}

\usepackage{amsfonts}
\usepackage{amssymb}
\usepackage{amsthm}
\usepackage{algorithm}
\usepackage{algorithmic}
\usepackage{booktabs}
\newcommand{\thetapre}{\theta_{\mathrm{pre}}}
\newcommand{\thetasft}{\theta_{0}}
\newcommand{\avg}{\mathrm{avg}}

\theoremstyle{plain}
\newtheorem{theorem}{Theorem}[section]
\newtheorem{proposition}[theorem]{Proposition}

\theoremstyle{definition}
\newtheorem{definition}[theorem]{Definition}
\newtheorem{assumption}[theorem]{Assumption}
\theoremstyle{remark}

\title{Mitigating the Alignment Tax of RLHF}

\author{
 \textbf{Yong Lin\textsuperscript{1}\thanks{indicates equal contributions, random order. Correspond to <hlinbh@connect.ust.hk>}},
 \textbf{Hangyu Lin\textsuperscript{2*}},
 \textbf{Wei Xiong\textsuperscript{3*}},
 \textbf{Shizhe Diao\textsuperscript{4*}},
 \textbf{Jianmeng Liu\textsuperscript{2}},
 \textbf{Jipeng Zhang\textsuperscript{2}},
 \textbf{Rui Pan\textsuperscript{3}},
 \\
 \textbf{Haoxiang Wang\textsuperscript{3}},
 \textbf{Wenbin Hu \textsuperscript{2}},
 \textbf{Hanning Zhang\textsuperscript{2}},
 \textbf{Hanze Dong\textsuperscript{2}},
 \textbf{Renjie Pi\textsuperscript{2}},
 \\
 \textbf{Han Zhao\textsuperscript{3}},
 \textbf{Nan Jiang\textsuperscript{3}},
 \textbf{Heng Ji\textsuperscript{3}},
 \textbf{Yuan Yao\textsuperscript{2}},
 \textbf{Tong Zhang\textsuperscript{3}}
\\
\\
  \textsuperscript{1} Princeton University, Princeton Language and Intelligence
  \\
 \textsuperscript{2}The Hong Kong University of Science and Technology
 \\
 \textsuperscript{3}University of Illinois Urbana-Champaign,
 \textsuperscript{4}NVIDIA
}

\begin{document}
\maketitle

\begin{abstract}
    LLMs acquire a wide range of abilities during pre-training, but aligning LLMs under Reinforcement Learning with Human Feedback (RLHF) can lead to forgetting pretrained abilities, which is also known as the alignment tax. To investigate alignment tax, we conducted experiments with existing RLHF algorithms using OpenLLaMA-3B, which revealed a pronounced alignment tax in NLP tasks. Whereas, despite various techniques to mitigate forgetting, they are often at odds with the RLHF performance, leading to a trade-off between alignment performance and forgetting mitigation, leading to an alignment-forgetting trade-off.
    
    In this paper we show that model averaging, which simply interpolates between pre and post RLHF model weights, surprisingly achieves the most strongest alignment-forgetting Pareto front among a wide range of competing methods. To understand its effectiveness, we offer theoretical insights into model averaging, revealing that it enhances performance Pareto front by increasing feature diversity on the layers where tasks share overlapped feature spaces. Empirical evidence corroborates our analysis by showing the benefits of averaging low-level transformer layers. Building on the analysis and the observation that averaging different layers of the transformer leads to significantly different alignment-forgetting trade-offs, we propose Heterogeneous Model Averaging (HMA) to Heterogeneously find various combination ratios of model layers. HMA seeks to maximize the alignment performance while incurring minimal alignment tax. Moreover, we validate HMA's performance across a range of RLHF algorithms over OpenLLaMA-3B and further extend our findings to Mistral-7B which is evaluated by open-sourced preference model and GPT4. Code available here\footnote{\url{https://github.com/avalonstrel/Mitigating-the-Alignment-Tax-of-RLHF.git}}.
    \vspace{-0.5em}
\end{abstract}

\section{Introduction}
\label{sect:intro}
Large Language Models (LLMs), such as GPT4 \citep{OpenAI2023GPT4TR}, Bard \citep{google@bard}, and Claude \citep{Anthropic@claude}, have attracted widespread attention due to their remarkable achievements. LLMs are pre-trained on vast datasets, which equip them with the ability to effectively handle diverse tasks, e.g., GPT-3 showcases its prowess in various tasks such as reasoning, common sense question-answering (QA), translation, and so on.

\begin{figure}
    \centering
    \includegraphics[width=0.8\linewidth]{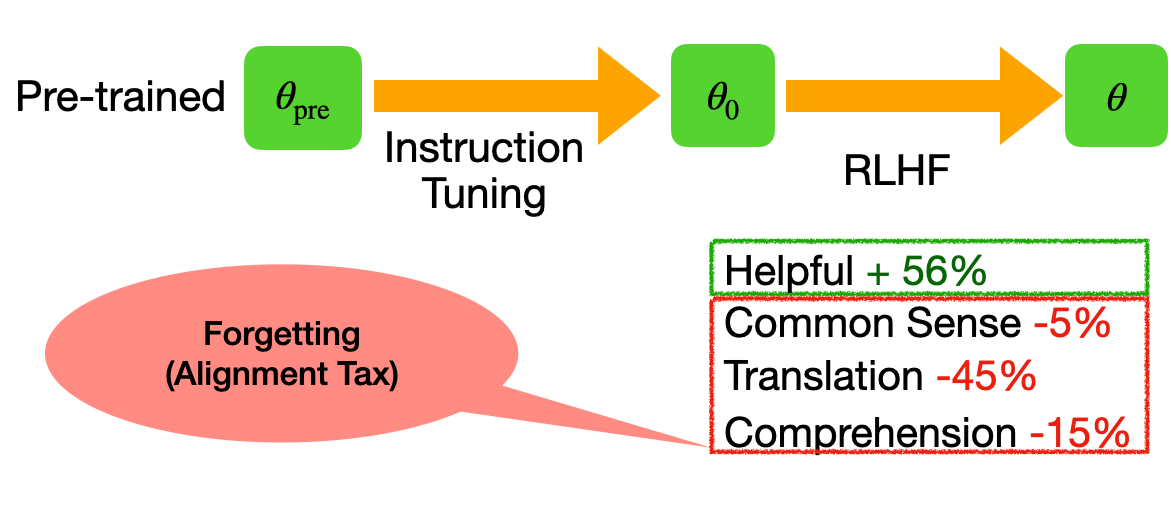}
    \caption{Illustration of RLHF procedure and the alignment tax.}
    \label{fig:RLHF_procedure}
    \vspace{-1.5em}
\end{figure}

While LLMs exhibit strong abilities among various benchmarks, they still require alignment with human preferences, including the principles of being helpful, honest, and harmless as outlined by \citep{askell2021general}.
The goal is to ensure that LLMs are designed to assist users in completing tasks, provide truthful information without deception, and avoid causing harm, whether physical, psychological, or social, to individuals or the environment. The process of aligning LLMs with human preferences often involves the application of Reinforcement Learning with Human Feedback (RLHF)~\citep{ouyang2022training}, as shown in Figure~\ref{fig:RLHF_procedure}.
Although RLHF allows LLMs to align with human expectations, prior studies~\citep{askell2021general, OpenAI2023GPT4TR, song2023reward} have found that this approach can lead to forgetting in the diverse abilities that the LLMs have already acquired, as illustrated in Figure~\ref{fig:RLHF_procedure}.
This phenomenon, also known as the ``alignment tax" in the literature, has accumulated substantial attention from both academia and industry \citep{ouyang2022training, Anthropic@claude, askell2021general, tu2023sight, noukhovitch2023language}.

\textbf{Investigating alignment tax}. In this paper, we first conduct a comprehensive investigation on alignment tax and develop methods to reduce alignment tax while maintaining the alignment performance. In particular, we followed the approach presented by~\citep{ouyang2022training} and evaluated alignment tax using multiple NLP benchmarks from common sense QA, such as ARC Easy and Challenge~\citep{clark2018arc}, Race \citep{lai2017race}, and PIQA \citep{bisk2020piqa}, reading comprehension benchmarks including SQuAD \citep{rajpurkar2018know} and DROP \citep{dua2019drop}, and translation tasks, including WMT 2014 French to English translation \citep{bojar2014findings} (c.f.~Section~\ref{sect:basic_settings}). Our primary focus is on aligning the OpenLLaMA-3B on the helpfulness and harmlessness dataset \citep{bai2022training} using Rejection Sampling Fine-tuning methods \citep{dong2023raft} (also known as the best-of-$n$ algorithm). In the later part, we extend our experiments to Mistral-7B and Direct Preference Optimization (DPO, \cite{rafailov2023direct}). We mainly focus on RSF and DPO since they are popular and nearly all of the latest open-sourced LLMs on the leaderboards are aligned by these two methods{\footnote{\scriptsize \url{https://tatsu-lab.github.io/alpaca_eval/}}}.  Indeed, we observed a substantial alignment tax on these benchmarks consistently, confirming the findings of \citep{ouyang2022training, gao2023scaling}. Specifically, as we gained a higher reward during RLHF, indicating better alignment with human preference, the alignment tax also increased simultaneously, clearly inducing a \textit{alignment-forgetting trade-off}.

\textbf{Surprising effectiveness of model averaging over}. We then compare various methods developed in different communities as potential rescues to alleviate the alignment tax. This includes the model averaging method \citep{wortsman2022robust, wortsman2022model, lin2023spurious} from out-of-distribution (OOD) generalization literature, regularization-based techniques from the continual learning literature \citep{panigrahi2023task, xuhong2018explicit, buzzega2020dark, huang2021continual}, low-rank adaptation (LoRA) \citep{huang2021continual} from the parameter-efficient fine-tuning literature, as well as the utilization of reward penalty from the reinforcement learning literature \citep{ziegler2019fine, wu2021recursively, ouyang2022training, yuan2023rrhf}. Interestingly, we found that model averaging, which simply interpolates between the weights of models before and after RLHF, achieves the most efficient alignment-forgetting Pareto front. In Appendix~\ref{app:experience_replay}, we further show and discuss the in-effectiveness of Experience Reply \cite{rebuffiincremental} method compared with MA.

\textbf{Understanding the effectiveness of model averaging}. To understand the effectiveness of model averaging, we provide theoretical insights based on the framework of~\citep{lin2023spurious}. In particular, we show that the method can enhance Pareto front by increasing feature diversity on layers where two tasks share similar feature spaces. Empirical evidence also indicates that averaging the low-level layers of Transformers consistently improves both alignment reward and NLP task performance. This aligns with our theoretical insights, as tasks could share similar lower-level features, e.g., better word representation on low-level layers benefits both NLP and alignment tasks.

\textbf{Heterogeneous model averaging}. We noticed that averaging different layers of the Transformers unveiled notably distinct patterns of alignment-forgetting trade-off, aligning with our earlier analysis that tasks may exhibit varying overlapping feature spaces in different layers. Motivated by this observation, we propose Heterogeneous Model Averaging (HMA), which adaptively averages different parts of the models during model averaging. We start by dividing the transformer into $K$ parts and assigning unique averaging ratios for each part, represented as $\alpha_i \in [0, 1]$ for the $i$th part. HMA aims to maximize alignment reward by optimizing the averaging ratios $(\alpha_1, \ldots, \alpha_K)$ while maintaining the overall alignment tax, thus consistently improve the alignment-forgetting Pareto front.
 To demonstrate the efficiency of HMA, we also contrasted our method with other RLHF techniques, including Direct Preference Optimization (DPO). \citep{rafailov2023direct} 
 We further substantiate our findings on Mistral-7B where evaluations conducted by open sourced perference model and GPT4, which further corroborates our empirical findings on OpenLLaMA-3B.
 
We summarize our contributions as follows:
\vspace{-.3em}
\begin{itemize}
    \item We provide a comprehensive investigation of the alignment tax challenge in RLHF on NLP tasks. We systematically compare a wide range of methods to alleviate alignment tax and highlight model averaging as a particularly effective approach.
    \item We provide theoretical insights into the efficiency of model averaging in enhancing the alignment-forgetting trade-off, demonstrating that both NLP and alignment tasks can benefit from the increased feature diversity from model averaging in the shared feature space.
    \item Motivated by our analysis, we introduce Heterogeneous Model Averaging (HMA), which optimizes the averaging ratios of different model layers to maximize alignment performance. HMA consistently improves the Pareto front across different benchmarks, and it also generalizes well across various RLHF algorithms and different model types, such as OpenLLaMA-3B and Mistral-7B, evaluated by open-sourced preference model and GPT4.
\end{itemize}

The paper is structured as follows: we conduct a systematic investigation of existing methods in Section~\ref{sect:basic_settings}-\ref{sect:results_existing}. In Section~\ref{sect:insights}, we provide insights into the effectiveness of model averaging. Subsequently, we propose Heterogeneous Model Averaging in Section~\ref{sect:HMA}. We conclude the paper in Section~\ref{sect:conclusion}.

\section{Discussion with existing works.} 
In this section, we provide comparison of this work with existing works to highlight the novelty of our findings. We defer more comprehensive related works to Appendix~\ref{app:related_work}. 

\textbf{Existing works of model averaging for LLMs.} 
Previous research has covered certain aspects of model averaging. \cite{rame2024warm} demonstrate the utilization of model averaging to construct a more resilient reward model for reinforcement learning with human feedback (RLHF). In a similar vein, \cite{rame2024rewarded}  employ model averaging to merge policy models trained for distinct objectives, facilitating multi-objective RLHF. \cite{sanyal2023early} introduce the integration of moving averaging to enhance pre-training. However, none of these studies investigate the alignment tax, and their findings are independent of our research. 

\textbf{Existing works on finding adaptive combinations for model merging.}
Previous studies \cite{yang2023adamerging, akiba2024evolutionary} have also discussed the idea of dynamically assigning different weights to different layers when merging models, aiming to maximize performance on a specific task (e.g., $\cT_i$). These approaches assume access to the task-specific data $\cT_i$. However, considering the nature of alleviating alignment tax, which aims to mitigate forgetting across a extremely wide range of tasks ($\cT_{j_1}...\cT_{j_K}$), these methods fail to effectively optimize performance for multiple tasks simultaneously. In the Appendix~\ref{app:ada_merging}, we demonstrate that using the method proposed by \cite{yang2023adamerging}, which optimizes for a single task, does not effectively address forgetting on the other tasks. Furthermore, our work is the first to provide an explanation for the surprising effectiveness of model averaging in alleviating forgetting, as well why we should assign heterogeneous combination ratios.


\textbf{Existing works on the forgetting of language models.} Most research on forgetting in language models focuses on sequentially pre-training \cite{chen2023lifelong, gong2022continual, jin2021lifelong, qin2022elle, liu2021continual} or fine-tuning tasks  \cite{sun2019lamol, razdaibiedina2023progressive, wu2021pretrained, zhang2022continual, madotto2020continual}, e.g., sequentially training on task $\cT_i$ and then task $\cT_j$. They evaluate forgetting by measuring the model's performance on a task 
(e.g., task $\cT_i$) after training it on another task (e.g., task $\cT_j$). However, these methods have not explored the effectiveness of model averaging. In our case, we demonstrate the significant power of model averaging which outperform a wide range of existing methods. Furthermore, existing works assume that the data size of each task is comparable (i.e., the dataset size of $\cT_i$ and $\cT_j$ is similar), allowing for a subset (e.g., 10\%) of old task data replay, which is shown to effective alleviate the forgetting without excessive computation overhead in their settings. However, in our alignment tax situation, we aim to preserve a wide range of abilities gained during pre-training, which is challenging since pre-training datasets are often not publicly available. In Appendix~\ref{app:experience_replay}, we show that even when we have access to the pre-training data and replay a subset up to four times larger than the RLHF data (which costs significant computation overhead), experience replay still under-performs model averaging in two out of three benchmarks. This is likely due to the vast size of the pre-training data, where the subset only covers a small fraction of it (e.g., only covers \textasciitilde $0.01\%$ of the pre-training data). So replay methods are less practical for alleviating alignment tax.

\vspace{-.5em}
\section{Experimental Settings}
\label{sect:basic_settings}
\textbf{Basic Setting.}~We chose the OpenLLaMA-3B model \citep{openlm2023openllama} because (1) it is computational friendly compared with 7B models (2) it has openly available pre-training dataset, which is convenient to investigate Experience Replay in Appendix.~\ref{app:experience_replay}. Furthermore, we extend the experiments to Mistral-7B in Sec.~\ref{sect:HMA}. Following the standard procedure outlined in \citep{ouyang2022training}, we initially conducted instruction tuning, followed by RLHF. Here, $\theta$ represents an LLM with parameters $\theta$, with the pre-trained model denoted as $\thetapre$. We commenced with instruction fine-tuning for $\thetapre$ on ShareGPT \footnote{\scriptsize \url{https://huggingface.co/datasets/anon8231489123/ShareGPT_Vicuna_unfiltered}}, which yielded $\thetasft$. Subsequently, RLHF was performed on $\thetasft$ to obtain $\theta$. Similar to the methodology proposed in \citep{ouyang2022training}, the alignment tax was evaluated by comparing the performance regression of $\theta$ with $\thetasft$ across various NLP tasks. The whole procedure and notations are illustrated in Fig.~\ref{fig:RLHF_procedure}. \looseness=-1



\textbf{Datasets for Evaluating Alignment Tax.}~Following the approach in \citep{ouyang2022training}, our evaluation of alignment tax encompasses various NLP benchmarks: (a) Common Sense QA: This includes ARC Easy and Challenge \citep{clark2018arc}, Race \citep{lai2017race}, and PIQA \citep{bisk2020piqa}, with the performance being assessed using accuracy. (b) Reading Comprehension:  we employ SQuAD \citep{rajpurkar2018know} and DROP \citep{dua2019drop} to gauge reading comprehension ability, with evaluation based on the F1 score for both datasets. (c) Translation: Our evaluation utilizes WMT 2014 French to English translation \citep{bojar2014findings}, with performance measured using BLEU \citep{papineni2002bleu} scoring.

\textbf{RLHF Basics.}
In our notation, $\pi_\theta$ denotes the policy induced by the LLM $\theta$.
Additionally, $x$ represents the input prompt and $a$ denotes the output (which is also referred to as an action in RL literature \citep{schulman2017proximal}). 
Drawing from \citep{ouyang2022training, bai2022training, dong2023raft, touvron2023llama, rafailov2023direct}, we assume the existence of a ground-truth reward function $r^*(x,a): \cX\times\cA \to [0,1]$, where $\cX$ and $\cA$ denote the spaces of $x$ and $a$ respectively. The primary objective of RLHF is to maximize:
\begin{align}
    \label{eqn:RLHF_basic}
    \max_{\theta} \bbE_x \bbE_{a \sim \pi_\theta (\cdot|x)} [r^* (x, a)].
\end{align}
\textbf{RLHF Algorithm.}~We adopt {Rejection Sampling Finetuning} (RSF) for our main experiments \citep{dong2023raft, touvron2023llama, yuan2023rrhf, gulcehre2023reinforced} and also further verify our findings on Proximal Policy Optimization (PPO) \citep{schulman2017proximal} and Direct Preference Optimization (DPO) \citep{rafailov2023direct} in Sec.~\ref{sect:HMA}. Essentially, the RSF learns from the best-of-n policy \citep{nakano2021webgpt}, which samples $n$ responses for each prompt query and returns the one with the highest reward. As suggested by \citep{dong2023raft, touvron2023llama, gulcehre2023reinforced}, we adopt an iterative training set-up for the implementation instead of always sampling the samples from the starting checkpoint because we find that the iterative training is far more sample-efficient. Specifically, for each iteration, we first sample a batch of prompts and generate $n$ responses for each prompt from the current model. Then, we use the reward model to compute the rewards for each prompt-response pair, and for each prompt, we select the one with the highest reward into a small subset. By this process, we collect a batch of samples from the best-of-n policy that are with high reward. We simply fine-tune the current model on this subset to get the next model and the next iteration begins.


\begin{figure}[t!]
    \centering
    \includegraphics[width=\linewidth]{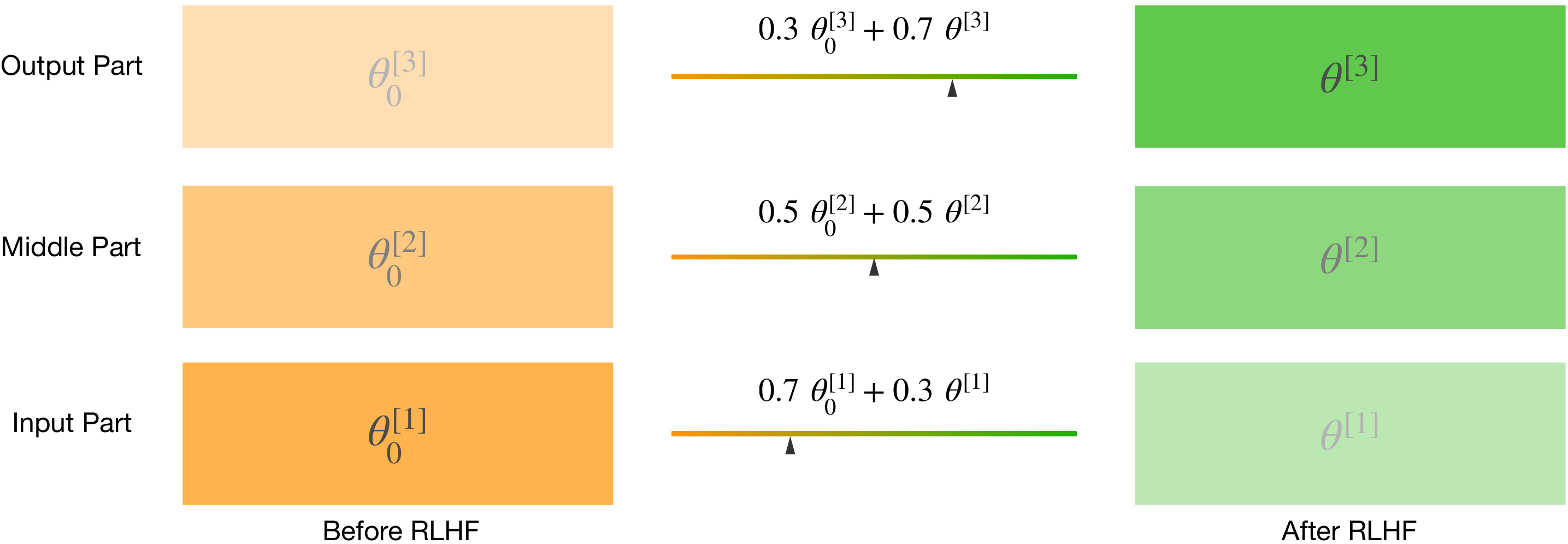}
    \vspace{-.8em}
    \caption{Illustration of Heterogeneous Model Averaging (HMA) when $K=3$.}
    \label{fig:HMA-demo}
    \vspace{-1em}
\end{figure}

\vspace{-.5em}

\section{Evaluating Existing Methods}
\label{sect:results_existing}
In Figure~\ref{fig:reward_tax_early_stopping} of Appendix~\ref{app:alignmnet_tax_during_training}, we visualize the training procedure in terms of the alignment-forgetting trade-off during RLHF. Specifically, we can clearly see that as the RLHF proceeds, the reward begins to increase while the translation and reading comprehension ability continues to drop.  Interestingly, we observe that the performance of common sense increases first and then drops. Given that alignment tax is inherently a catastrophic forgetting issue, we then proceed to explore methods to reduce alignment tax. Research focused on reducing forgetting is mainly classified into two main categories, depending on the availability of the pre-training dataset. We also investigate the reward penalty method developed in RL community in Appendix~\ref{app:reward_penalty}. 

\subsection{Basic Methods}
To explore methods for alleviating alignment tax, we initially examine solutions that do not rely on pre-training datasets. These methods encompass the following:(a) Early stopping. (b) Regularization towards $\thetasft$ in the weight space as follows:
\begin{align}
    \max_{\theta} \bbE_x \bbE_{a \sim \pi_\theta (\cdot|x)} [r^* (x, a)] + \lambda \|\theta - \thetasft\|_\alpha,
\end{align}
where we use $\alpha=1, 2$ which corresponds to the L1 and L2 \citep{xuhong2018explicit} penalties, respectively.
(c) Low-Rank Adaptation (LoRA) \citep{Hu2021LoRALA}. It introduces trainable rank decomposition matrices into linear layers to update $\theta - \thetasft$ during RLHF. (d) Knowledge distillation \citep{buzzega2020dark, huang2021continual}. We use $\pi_{\thetasft}$ serves as the teacher and $\pi_{\theta}$ as the student, with a penalty imposed as:
          \begin{align*}
              \max_{\theta} \bbE_x \bbE_{a \sim \pi_\theta (\cdot|x)} [r^* (x, a)] + \lambda \|\pi_{\theta}(\bx) -  \pi_{\thetasft}(\bx) \|_2^2.
          \end{align*}
(e) Model Averaging (MA) \citep{wortsman2022model,wortsman2022robust}.  This involves simply interpolating between  $\thetasft$ and $\theta$ to yield the policy $\pi_{(1-\alpha) \thetasft + \alpha \theta}$, where $\alpha$ is a hyper-parameter ranging from 0 to 1. (f) Stochastic Moving Averaging (SMA) \cite{noukhovitch2024language}. More implementation details are provided in the appendix.

\textbf{Results.}~Figure~\ref{fig:no_data_baseline} depicts the performance of each aforementioned method. The results demonstrate that these approaches effectively alleviate the alignment tax; however, they also result in a reduction in the RLHF reward, indicating a clear trade-off between reward and alignment tax. Notably, despite its simplicity, the Pareto-front of model averaging supersedes nearly all other methods across various hyper-parameters. In Appendix~\ref{app:experience_replay} and \ref{app:reward_penalty}, we compared model averaging with Experience Replay (ER) and KL reward penalty methods for Proximal policy optimization \cite{schulman2017proximal} algorithms, the conclusions are similar.  
\vspace{-.5em}

\begin{figure*}
    \centering
    \includegraphics[width=0.3\linewidth]{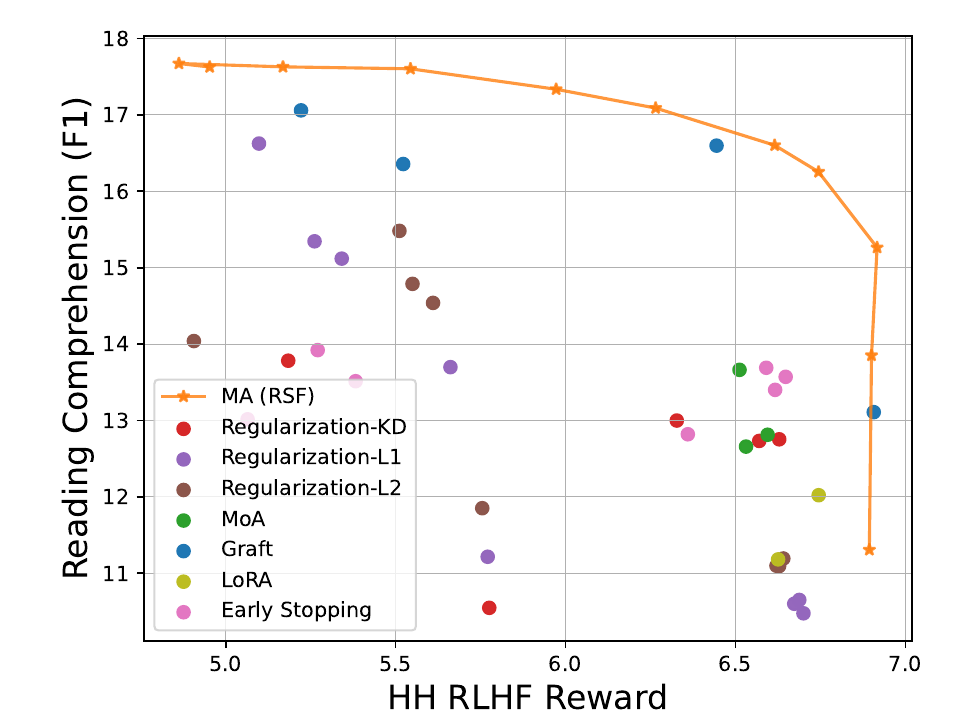}
    \includegraphics[width=0.3\linewidth]{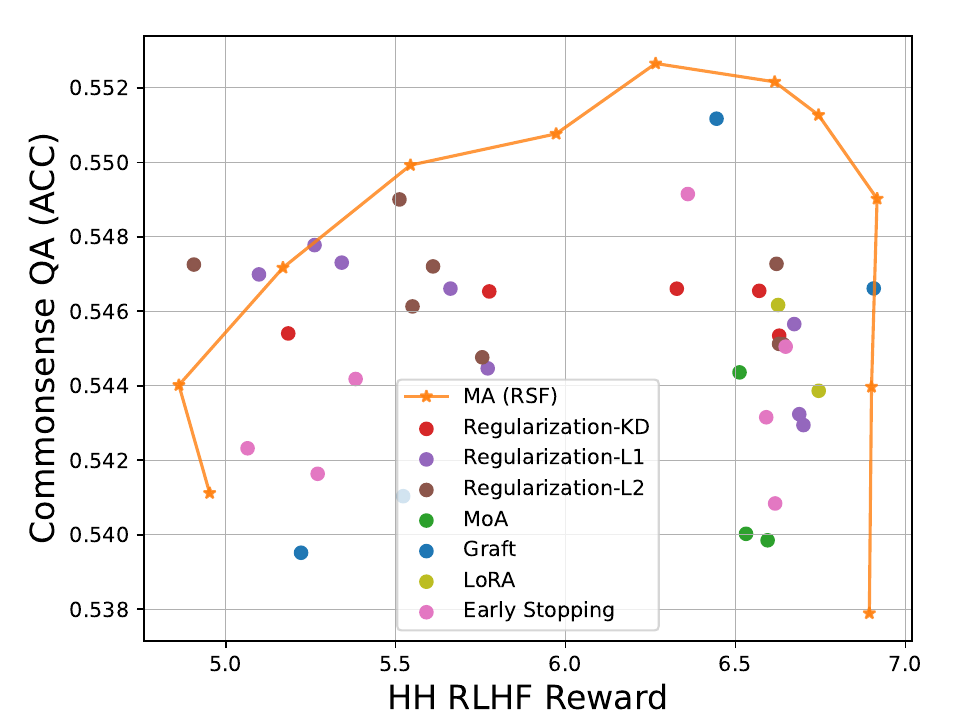}
    \includegraphics[width=0.3\linewidth]{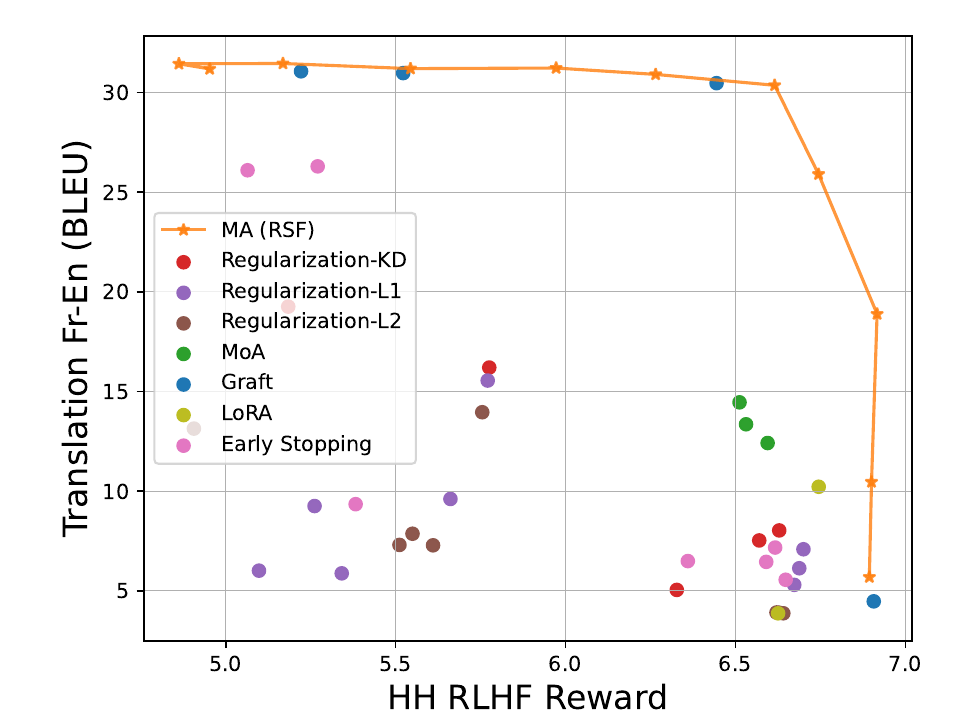}
    \vspace{-1em}
    \caption{Existing methods without access to pre-training data}
    \label{fig:no_data_baseline}
    \vspace{-1em}
\end{figure*}

\section{Unravelling the Mysteries of Model Averaging for Alleviating Alignment Tax}
\label{sect:insights}
Given the promising performance of model averaging, we try to understand the efficacy of model averaging in this Section and motivate our method to improve it. We utilize the theoretical framework proposed by \citep{lin2023spurious} to gain insights into its effectiveness in alignment tax. While the framework addresses classification problems, the insights derived can aid our understanding of model averaging. We also conduct empirical analysis using a generative model (Openllama-3B) to verify these theoretical insights. Analyzing the performance of model averaging in alignment tax is more intricate compared to the work of the study by \citep{lin2023spurious} focuses on out-of-distribution (OOD) scenarios, where the same task is performed under different distributions. In contrast, our focus in alignment tax is to comprehend the performance trade-offs among different tasks. To illustrate, consider the entire feature space $\cY$ and two tasks with label spaces $\cY_a \subset \cY$ and $\cY_b \subset \cY$, with the simplifying assumption that $|\cY_a|= |\cY_b| = K$. While \citep{lin2023spurious} only considers the case where $\cY_a = \cY_b$, we extend these results to encompass the case where $\cY_a \neq \cY_b$.

\textbf{Theoretical Settings.} Suppose we have many features $\cS_x = \{\bx_i\}_{i=1}^D$ where each feature $\bx_i \in \bbR^d$ and the observed feature $\bx \in \bbR^{d \times D}$ is a concatenation of $\bx_1, ..., \bx_D$. Following \citep{lin2023spurious}, we adopt a simplified model $f(\bx) = w \Phi(\bx)$ where $w \in \bbR^{d \times K}$, $\Phi(\bx) = \sum_{i=1}^D \Phi_i \bx_i$ and $\Phi_i \in \{0, 1\}, \forall i$.  Suppose we have two models $f_a(\cdot) = w_a \Phi_a(\cdot)$ and $f_b = w_b \Phi_b(\cdot)$ for tasks $\cT_a$ and $\cT_b$, respectively, relying on feature sets $\cS_{x, a} \subset \cS_x$ and $\cS_{x, b}  \subset \cS_x$, with $|\cS_{x, a}|=|\cS_{x, b}|=n$, and $|\cS_{x, a} \cap \cS_{x, a}| = n_o$ overlapped features. The averaged model of $f_a$ and $f_b$ is $f_{\avg}(\cdot) = w_{\avg} \Phi_{\avg}(\cdot)$, where $w_{\avg} = (w_a + w_b)/2$ and $\Phi_{\avg, i} = (\Phi_{a, i} + \Phi_{b, i}) / 2, \forall i$ \citep{lin2023spurious}.
To gain an intuitive understanding, we compare model averaging in two cases: Case (1) when the tasks are quite similar ($|\cY_A \cap \cY_B| = K$) and Case (2) when the tasks are independent ($|\cY_A \cap \cY_B| = 0$). \footnote{Notably, the overlap in features is independent of the overlap in label space. For instance, when classifying a dog, we can use either the animal shape or the texture (overlapped label space, non-overlapped feature); when classifying a dog or a cat, we can both use the animal shape (non-overlapped label space, overlapped feature).} Furthermore, even if the tasks are very similar, fitting two models on them can rely on different features due to randomness in data or training procedures \citep{lin2023spurious, allen2020towards}. We will investigate the performance of model averaging in Case (1) and (2) to gain insights on when it works. Following \citep{lin2023spurious}, we assume each feature is weak, failing with probability $p$. The effectiveness of model averaging is given by 
$$\xi = \frac{1}{2}\left(\cA_a(f_{\avg}) -  \cA_a(f_a) + \cA_b(f_{\avg}) -  \cA_b(f_b)\right),$$ 
where $\cA_a(f)$ and $\cA_b(f)$ denote the accuracy of $f$ on task $a$ and $b$, respectively.  We use $\xi^{(1)}$ to denote the effective averaging robustness for Case (1) and similarly define $\xi^{(2)}$ for Case (2).

\begin{proposition}
    \label{prop:intuitive_analysis}
    Consider the assumptions specified in the appendix. We have:
    \begin{align*}
        \xi^{(1)} - \xi^{(2)}  = & F_p \left( \frac{\sqrt{2}(1-p) n}{\sqrt{n+n_o}} \right) \\
        &-F_p \left((1-p)\sqrt{n} \right) \geq 0,
    \end{align*}
    where the equality holds when $n_o=n$ and $F_p(x)$ is a cumulative density function in Appendix~\ref{app:fp}.
\end{proposition}

\paragraph{Implications.} Proposition~\ref{prop:intuitive_analysis} demonstrates that when $\cT_a$ and $\cT_b$ are more similar, the averaging of models ($f_a$ and $f_b$) yields greater improvement. However, this improvement is reduced if $f_a$ and $f_b$ use more overlapping features. Recall that each weak feature can fail with probability $p$. If $\cT_a$ and $\cT_b$ are similar, the feature utilized by the two models would be projected into a shared space, allowing model averaging to take advantage of a more diverse set of features. This diversity reduces the probability of model failure because a diverse set of features is less likely to fail together simultaneously \citep{lin2023spurious}. However, if $\cT_a$ and $\cT_b$ are dissimilar, for example, if $|\cY_a \cap \cY_b|=0$ and the feature spaces corresponding to $\cY_a$ and $\cY_b$ are disjoint, then the features in the space of $\cY_a$ would not provide any information for predicting $\cY_b$. Therefore, averaging $f_a$ and $f_b$ would not improve the prediction of either task in this case. Refer to Appendix~\ref{app:theory_discuss} for a detailed discussion.

Notably, the model $\thetasft$ excels in NLP abilities before RLHF, while the model $\theta$ excels in alignment reward after RLHF. Using an analogy, we can equate NLP tasks with $\cT_a$, alignment with $\cT_b$, $\thetasft$ to $f_a$, and $\theta$ to $f_b$.  Recall that we adopt a simplified model for theoretical analysis by considering only one layer feature learner, although, in practice, we average a deep Transformer with 26 layers. Research has shown that different layers in deep neural networks capture varying levels of features \citep{yosinski2015understanding, zeiler2014visualizing, simonyan2014very}. For instance, low-level layers capture low-level features. Furthermore, tasks share similar feature space at a low level (alternatively, from the perspective of low-level layers, tasks look more similar). For example, improving the low-level features such as better word representation could enhance both RLHF reward and NLP tasks. Therefore, according to Proposition~\ref{prop:intuitive_analysis}, averaging the low-level layers could potentially elicit more improvements in both $\cT_a$ (NLP tasks) and $\cT_b$ (alignment reward) than higher layers.

\begin{figure}
\centering
\includegraphics[width=0.38\linewidth]{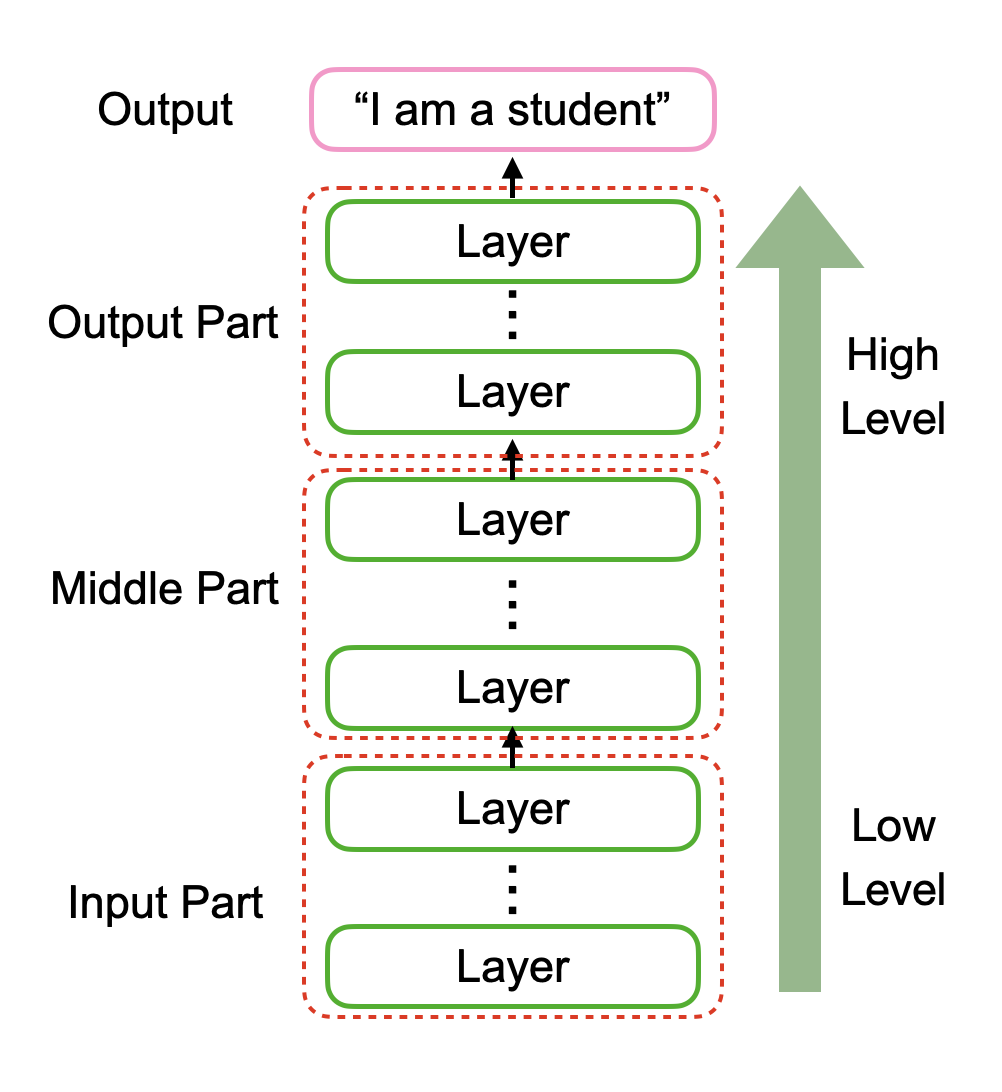}
\includegraphics[width=0.6\linewidth]{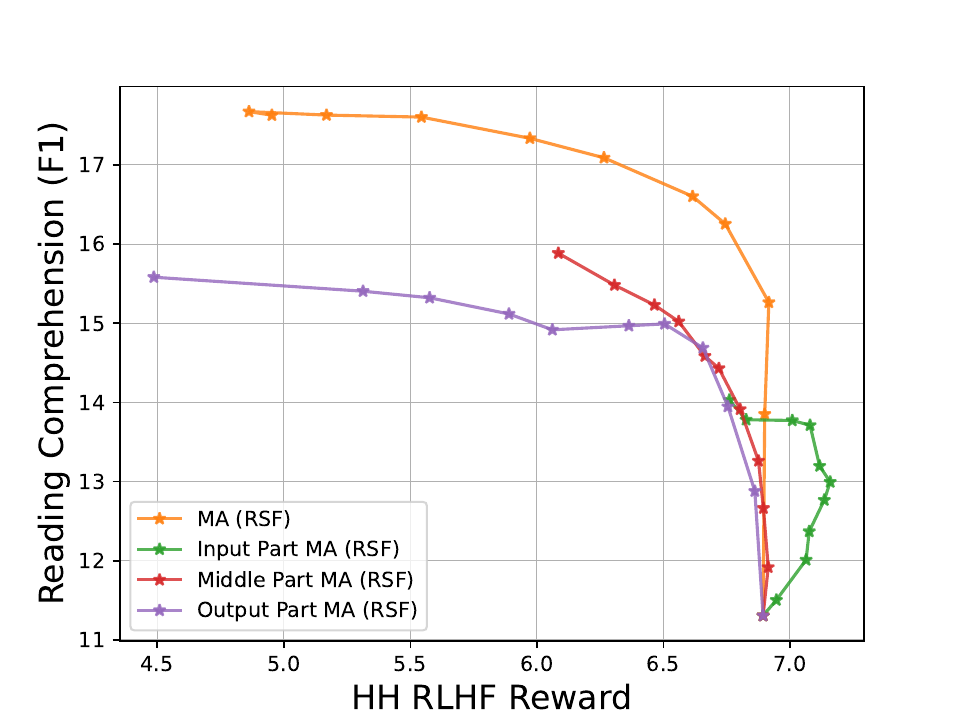}
\vspace{-2em}

\caption{(Left) Illustration of proof of concept experiments. We divide the Transformer into 3 parts. We only average one part each time. (Right) Merging different parts of the transformers.}
\label{fig:layerwise}
\label{fig:part_merging}

\end{figure}

\textbf{Empirical Validation.}~We categorize the 26 transformer layers of Openllama into three parts: the input part (layers 1-8), the middle part (layers 9-17), and the output part (layers 18-26). This division is depicted in Figure~\ref{fig:layerwise}.
We use the superscripts $[1]$, $[2]$, and $[3]$ to denote the input, middle, and output parts, respectively. For instance, $\theta^{[2]}$ represents the middle layers (9-18) of $\theta$. Here, $\thetasft$ and $\theta$ respectively refer to the models before and after RLHF. We investigate the impact of averaging one part instead of the whole Transformer: given a combination ratio $\alpha \in [0, 1]$, we average the $i$-th part of $\theta$ (i.e., $\theta^{[i]}$) with the corresponding part of $\thetasft$ (i.e., $\thetasft^{[i]})$, while keeping the remaining two parts of $\theta$ unchanged. So when we average the input part, the $j$-th part of the averaged model is:
\begin{align*}
    j\mbox{th part} = \begin{cases}
                          \alpha\theta^{[j]} + (1-\alpha)\thetasft^{[j]}, \mbox{ if } j = 1, \\
                          \theta^{[j]}, \mbox{ if } j = 2, 3.
                      \end{cases}
\end{align*}
The results of the above scheme are denoted as ``Input Part MA". ``Middle Part MA" and ``Output Part MA" represent that we average the middle and output parts, respectively.  Figure \ref{fig:part_merging} illustrates that the alignment-forgetting trade-off varies distinctly when different parts of the transformers are averaged.  Specifically, when we average the low-level layers, we observe a ``magical'' improvement in both the NLP tasks and alignment rewards, which is consistent with our previous analysis. Furthermore, we show results in Appendix~\ref{app:more_ave_diff_parts} that the magical improvement in averaging the low-level parts is consistent among DPO and PPO models.

\vspace{-1em}
\section{Heterogeneous Model Averaging}
\label{sect:HMA}

We have already shown that averaging different layers results in diverse patterns of alignment-forgetting trade-off \citep{Wu2022PretrainedLM,Lee2022SurgicalFI}. Therefore, different layers should not be equally treated during averaging. This leads to a natural question: can we enhance the alignment-forgetting trade-off by using adaptive weights for different layers? Consequently, we conduct proof-of-concept experiments to provide affirmative answers to this question and subsequently propose a practical algorithm.


\textbf{Proof of Concept.}~The following proof of concept experiments provide insights into average different layers with various ratios. We use different averaging ratio, i.e., $\alpha_1, \alpha_2, \alpha_3$, for the three parts. Specifically, the $i$th part of the averaged model is simply $\alpha_i \theta^{[i]} +  (1 - \alpha_i) \thetasft^{[i]}$. We try three patterns experiment given a base $\alpha \in \{0.2, 0.3, 0.4\}$ :  (a) $\alpha_1 = \alpha_2 = \alpha_3 = \alpha$; (b) $\alpha_1 = \alpha_2 = \alpha$, $\alpha_3 = \alpha - 0.1$, and (c) $\alpha_1 = \alpha$, $\alpha_2 = \alpha_3 = \alpha - 0.1$. We use $(\alpha |  \alpha | \alpha)$, $(\alpha |  \alpha | \alpha-0.1)$ and $(\alpha | \alpha-0.1| \alpha-0.1)$  to denote these three patterns, respectively. These results confirm that certain ratio combinations exceed the trade-off curve of vanilla model averaging, as displayed in Figure~\ref{fig:proof_of_concept} in Appendix~\ref{app:proof_of_concept_combination}.
Notably, some combination ratios consistently outperform the equal ratio across various benchmarks. This affirms the potential to identify consistent combination ratios that demonstrate superior performance across a broad spectrum of benchmarks in terms of alignment-forgetting trade-off.

\textbf{Heterogeneous Model Averaging.}~Upon dividing the Transformer into $K$ parts, our objective is to adaptively determine a combination ratio for different layers that consistently perform well across an extensive range of tasks. The conventional averaging method uses a shared $\alpha$ for all layers, playing a crucial role in defining the trade-off between reward and tax. We aim to identify an optimized combination of $(\alpha_1, ..., \alpha_K)$ to replace a uniform $\alpha$. Let $\theta(K)$ represent the model merged by $(\alpha_1, ..., \alpha_K)$. In particular, the $k$th component of the merged model $\theta(K)$ is given by
\begin{align*}
    \theta^{[k]}(K) := \alpha_k \theta^{[k]} +  (1 - \alpha_k) \thetasft^{[k]}, \forall k \in 1,...,K.
\end{align*}
To optimize the Pareto-front influenced by $\alpha$, we identify combination ratios corresponding to each $\alpha$. Subsequently, we establish the mean of $(\alpha_1, ..., \alpha_K)$ as $\alpha$ and ascertain the best combination of $(\alpha_1, ..., \alpha_K)$ to maximize the reward. Specifically, denoting $\Omega:= \left\{\frac{1}{K} \sum_{k} \alpha_k = \alpha, \alpha_1, ..., \alpha_K \in [0, 1]\right\}$, we solve:
\begin{align}
    \max_{(\alpha_1, ..., \alpha_K) \in \Omega} \mathbb{E}_{x} \mathbb{E}_{a \sim \pi_{\theta(K)}(\cdot | x )} \left[ r^*(x, a)\right].
\label{eqn:adaptive_model_averaging}
\end{align}

The intuition behind HMA is outlined as follows: (1) When maintaining the mean, i.e., $\frac{1}{K} \sum_{k} \alpha_k$, as $\alpha$, we can compare HMA performance with the performance of vanilla model averaging with the same $\alpha$. (b) We only optimize $K$ parameters, where $K$ is typically small. For example, we adopt $K=3$ by default and also include results with varying $K$ to the ablation study. This helps to ensure that the forgetting level of $(\alpha_1, ..., \alpha_K)$ remains close to $\alpha$. Intuitively, if we optimize a large number of parameters, it could easily lead to over-fitting in the in-domain (RLHF reward) and may also result in more significant forgetting. The whole algorithm is summarized Algorithm \ref{algo:hma} in appendix.

\begin{figure*}
    \centering
    \includegraphics[width=0.28\linewidth]{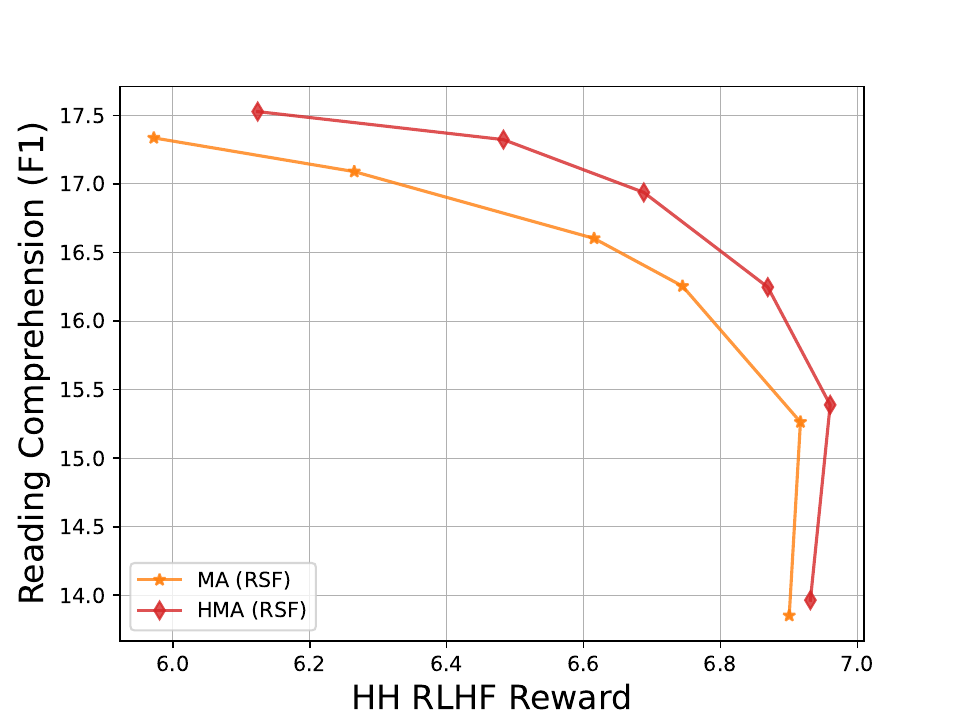}
    \includegraphics[width=0.28\linewidth]{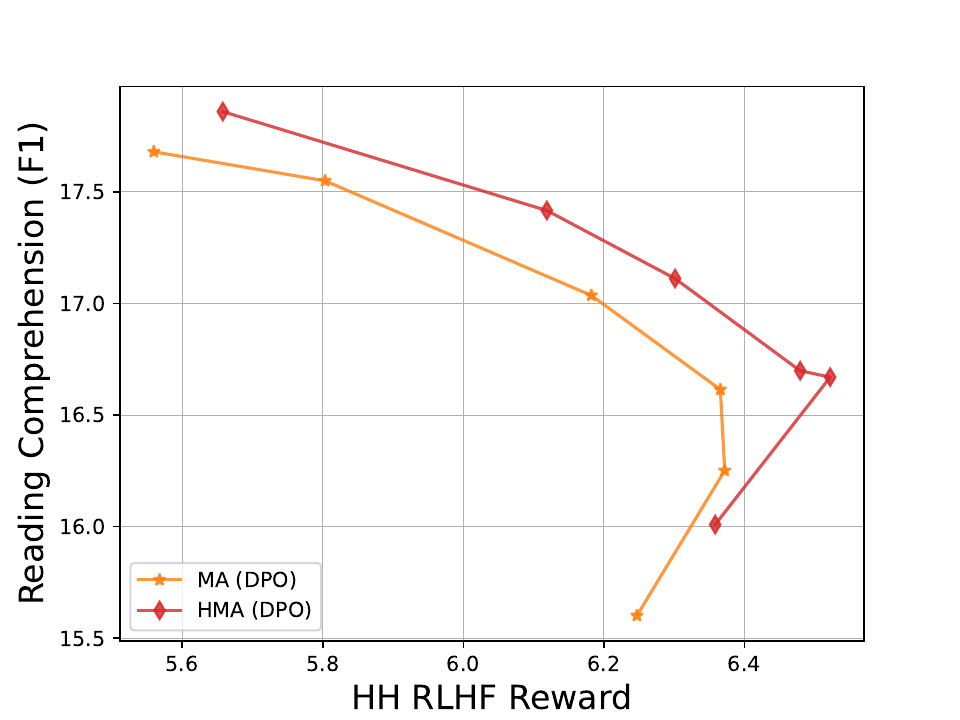}
    \includegraphics[width=0.28\linewidth]{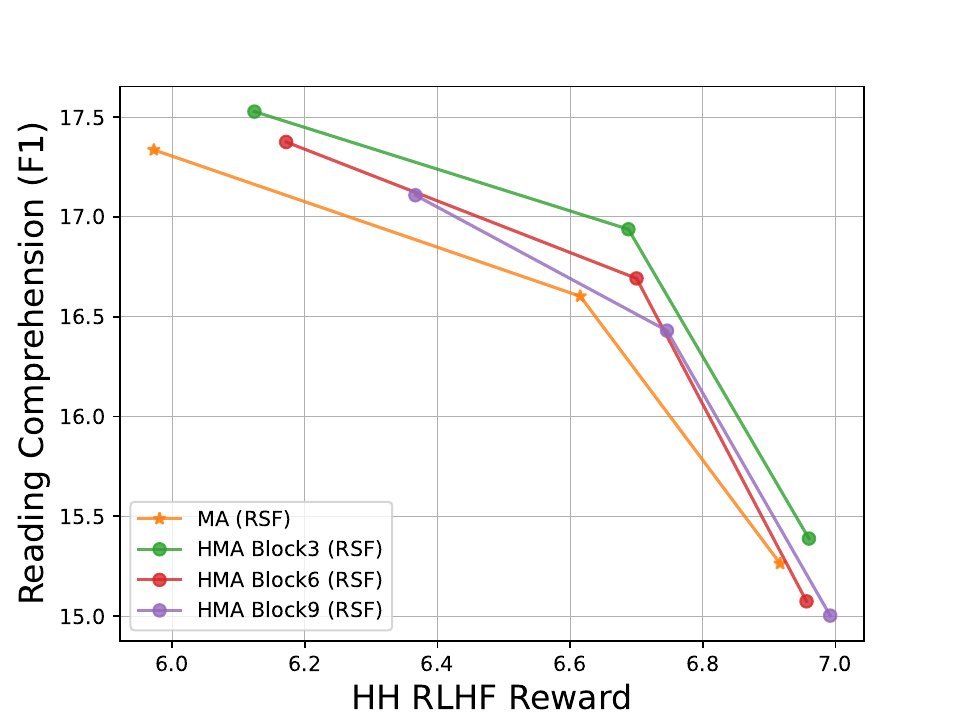}
    \vspace{-1em}
    \caption{Results of our HMA. (Top) HMA for RSF ( $\alpha \in [0.1, 0.6]$), (Bottom) HMA for DPO ( $\alpha \in [0.1, 0.6]$). (Right) HMA for RSF with different choices of $K$.  Refer to Appendix~\ref{app:detailed_HMA} for more results.}
    \label{fig:adaptive_model_averaging}
    \vspace{-1em}
\end{figure*}

\textbf{Results.}~The results of HMA are shown in Figure~\ref{fig:adaptive_model_averaging}. We can see that HMA can consistently push forward the Perato-front of the vanilla model averaging. Furthermore, such improvement is consistent over various RLHF algorithms.  More detailed results (e.g., on Commonsense QA and Translation with different RLHF algorithms) of HMA  can be found in Appendix~\ref{app:detailed_HMA}.

\textbf{Ablation results on different $K$.}~We tested different values of $K$ with $\alpha=0.2,0.4,0.6$ as illustrated in Figure~\ref{fig:adaptive_model_averaging} (Right). The trade-off curve shows a slight decrease as we increase $K$ from 3 to 6 and 9, but still consistently improves over the vanilla model averaging. This decrease is likely due to overfitting. Specifically, comparing the performance of HMA with different $K$ for the same mean ratio, we observe that as the alignment reward increases with an increase in $K$ from 3 to 9, the reading comprehension performance drops. 

\begin{figure}
    \vspace{-1em}

    \centering
     \includegraphics[width=0.6\linewidth]{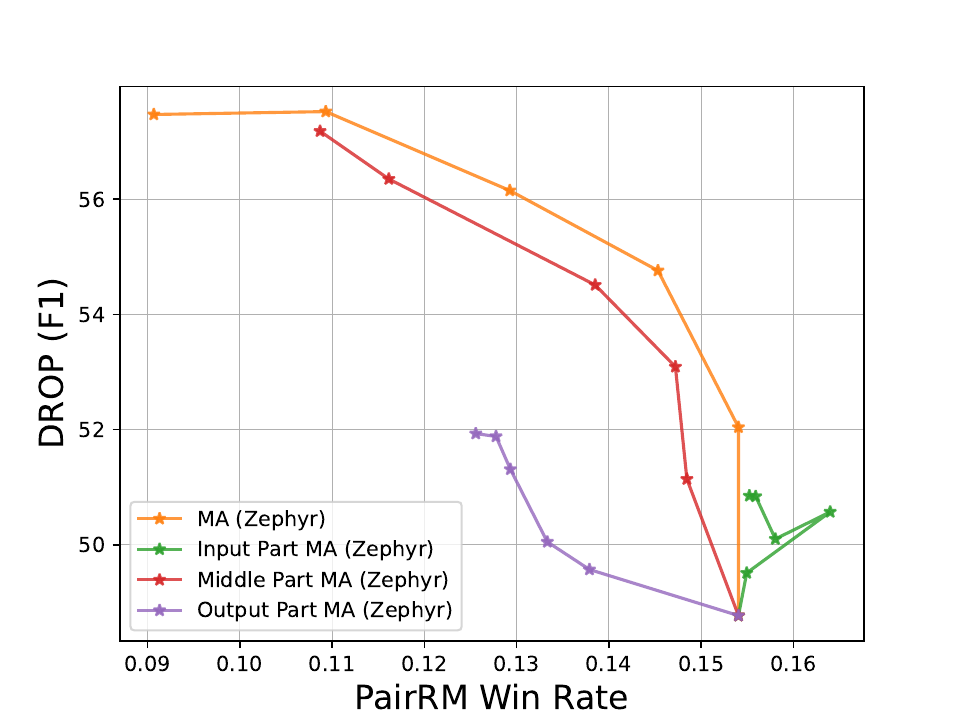}

   \includegraphics[width=0.6\linewidth]{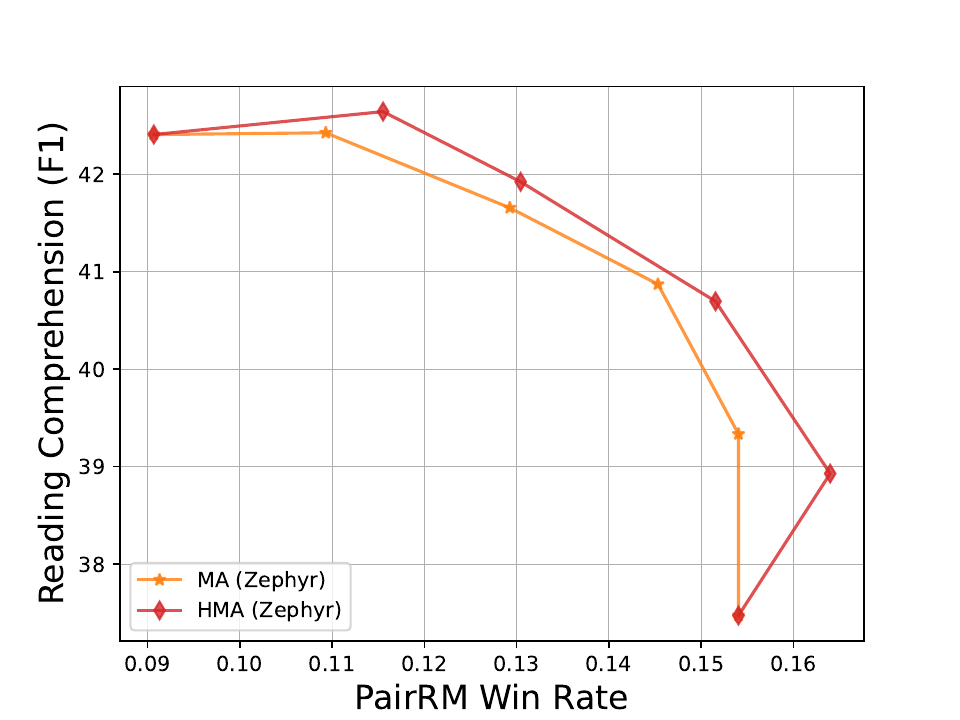}
 
    \caption{Results of Zephyr-7B-$\beta$ evaluated by open sourced preference model. (Top) Similar trends evaluated by PairRM when we average different blocks. (Bottom) Our HMA consistently improve over MA.} 
    \label{fig:res_zephyr_7b}
        \vspace{-1em}

\end{figure}

\textbf{How to choose the averaging ratio}. In practice, we determine the averaging ratio $\alpha$ for adopting vanilla MA or our HMA. Changing the averaging ratio for MA and HMA is convenient as these methods are applied after training the vanilla RLHF checkpoint. The comprehensive results in Figures~\ref{fig:no_data_baseline}, \ref{fig:adaptive_model_averaging}, and \ref{fig:detailed_results_ama} (details in Appendix~\ref{app:alpha_02}) show that $\alpha=0.2$ can consistently alleviate the alignment tax without hurting alignment performance. Further results of Zephyr-7B are shown in Figure~\ref{fig:res_zephyr_7b}. Additionally, the performance of the averaging ratio on different benchmarks (Figure~\ref{fig:proof_of_concept}) exhibits similar trends. Hence, we believe $\alpha=0.2$ is a suitable choice that can generalize to more tasks.

\begin{table}[htb!]
       \centering
    \resizebox{\linewidth}{!}{
    \begin{tabular}{c|c|ccc}
         Model & Win-Rate & Reading & CommonSense& Trans \\
         \midrule
         Zephyr-7B-$\beta$ & 8.10\% & 37.47 &  66.34 &36.55 \\
        HMA (Ours) & \textbf{9.32}\% & \textbf{38.93} & \textbf{66.55} & 
        \textbf{37.23} \\
        \midrule
        Zephyr-7B-Gemma & 11.3\% & 41.15 &  66.3 &38.09 \\
        HMA (Ours) & \textbf{11.5}\% & \textbf{42.45} & \textbf{66.4} & 
        \textbf{38.71}
    \end{tabular}
    }
    \caption{GPT4 evaluation of experiments of Zephyr-7B-$\beta$ and Zephyr-7B-Gemma on Alpaca benchmark. Reading is short for Reading Comprehension, which is evaluated by F1. CommonSence is evaluated by Accuracy (\%). Trans is short for Translation Fr-En, evaluated by BLEU.}
    \label{tab:GPT4}
    \vspace{-0.5em}
\end{table}

\textbf{Other models results.}~To further validate our method on larger LLMs, e.g., Mistral-7B \citep{jiang2023mistral} based models, we apply model averaging (MA) and Heterogeneous Model Averaging (HMA) on Zephyr-7B-$\beta$\footnote{\scriptsize \url{https://huggingface.co/HuggingFaceH4/zephyr-7b-beta}} \citep{tunstall2023zephyr} which is trained with DPO on the SFT version, Mistral-7B-SFT-$\beta$\footnote{\scriptsize \url{https://huggingface.co/HuggingFaceH4/mistral-7b-sft-beta}}. We also apply HMA on Zephyr-7B-Gemma  \footnote{\scriptsize \url{https://huggingface.co/HuggingFaceH4/zephyr-7b-gemma-v0.1}} which is aligned based on Gemma-7B\footnote{\scriptsize \url{https://huggingface.co/google/gemma-7b}} model.
Here we use the the publicly available preference model PairRM~\citep{jiang2023llm} to judge the helpfulness and evaluate models on AlpacaEval 2.0~\citep{alpaca_eval}. We reports the win rates of each model. Figure~\ref{fig:res_zephyr_7b}~(Top) shows that the trends of averaging different layers evaluated by PairRM are similar with the results evaluated by our own reward model.
The results range across $\alpha=0,0.2,\dots,1.0$ depicted in Figure~\ref{fig:res_zephyr_7b}~(Bottom) demonstrate that MA effectively achieves a strong Pareto front to mitigate forgetting in the Mistral-7B models. Additionally, our HMA algorithm shows further improvement compared to the MA method. 

\textbf{GPT4 Evaluation}. We also use GPT4 to evaluate HMA on AlpacaEval 2.0~\citep{alpaca_eval}. Due to the limited quota, we only compare HMA with $\alpha=0.2$ with vanilla Zephyr-7B-$\beta$ ($\alpha=0.2$ is recommended by the previous discussion). In Table~\ref{tab:GPT4}, we summarize their Win-Rate against GPT4 as well as their performance on NLP tasks. We show that HMA consistently outperforms Zephyr-7B-$\beta$ on all the metrics.

\section{Conclusion}
\label{sect:conclusion}
In this paper, we highlight the surprisingly effectiveness of model averaging and propose the Heterogeneous Model Averaging (HMA) framework to further enhance the performance.


\section*{Limitations}
Though our HMA significantly alleviates the alignment tax, it has not been fully eliminated. Future work could explore the theoretical lower bound of the alignment tax and determine which method could achieve the optimal trade-off.



{\small
\bibliography{main}
}

\newpage
\appendix
\onecolumn
\section{Related Work}
\label{app:related_work}
\paragraph{Large Language Models.} Large Language Models (LLMs) are pre-trained using vast amounts of data and has the ability to handle a diverse set of tasks.
An excellent line of LLMs includes  GPT \citep{brown2020language, OpenAI2023GPT4TR}, Bard \citep{google@bard}, Claude \citep{Anthropic@claude}, LLaMA \citep{touvron2023llama}, Galactica \citep{taylor2022galactica}, Bloom \citep{scao2022bloom}.
It is a common practice to fine-tune the LLMs to obtain better performance on a specific task~\citep{diao2023lmflow}, follow the instruction of humans \citep{ouyang2022training, sanh2021multitask, wang2022self} and align with humans' preferences \citep{christiano2017deep, askell2021general,bai2022training, ouyang2022training, dong2023raft}.

\paragraph{Reinforcement Learning with Human Preference (RLHF).}
RLHF \citep{christiano2017deep} has attracted considerable attention in the past few years, particularly after the tremendous success of the ChatGPT \citep{ouyang2022training, OpenAI2023GPT4TR}.
There is a rich literature on RLHF and the related discussions which cannot be comprehensively reviewed here due to the space constraint.
We thus refer the interested readers to the survey paper like \citep{casper2023open} but focus on the algorithmic designs here.
Proximal Policy Optimization (PPO) \citep{schulman2017proximal} is the predominant approach in RLHF whose effectiveness has been showcased by ChatGPT \citep{OpenAI2023GPT4TR}, Claude \citep{Anthropic@claude}, and Bard \citep{google@bard}.
However, it is known that the PPO is unstable and sample-inefficient in aligning LLMs \citep{choshen2019weaknesses} and imposes a heavy burden on the GPU resources as it requires loading multiple (typically four) models at the same time \citep{yuan2023rrhf, dong2023raft}.
In view of this, attempts have been made to propose alternative approaches to the PPO algorithm.
There is a line of work using the rejection sampling (also referred to as the best-of-$n$ sampling in the literature) \citep{nakano2021webgpt}, to reinforce the dataset used to finetune the LLM, including \citep{dong2023raft, yuan2023rrhf, touvron2023llama, gulcehre2023reinforced}.
Among them, \citep{dong2023raft, touvron2023llama, gulcehre2023reinforced} adopt an iterative framework, which is more sample-efficient and effective, while \citep{yuan2023rrhf} highlights the importance of sampling strategy.
In comparison to the original rejection sampling algorithm, which generates $n$ responses but only output the one with the highest reward, the LLMs aligned by iterative rejection sampling balance the goal of alignment and the inference cost. Meanwhile, there is also another line of work aiming to derive algorithms from the reverse KL-constrained contextual bandit \citep{rafailov2023direct, zhao2023slic, wang2023beyond, azar2023general, Xiong2023GibbsSF}, whose theoretical property is studied in \citep{Xiong2023GibbsSF}.
Among them, Direct Preference Optimization (DPO) \citep{rafailov2023direct} has appeared to be one of the most attractive algorithms, which optimizes the LLMs without the reward modeling and directly by preference learning from an offline dataset.
In view of the success of DPO, there has also been a debate on whether reward modeling is necessary, where \citep{rafailov2023direct, zhao2023slic, azar2023general} support bypassing reward modeling.
Although there are many works on reward optimization, the forgetting issue (also referred to as the alignment tax \citep{casper2023open} in the literature) of RLHF algorithms has not been comprehensively studied. Therefore, we choose three representative algorithms, including the PPO \citep{schulman2017proximal}, RSF \citep{dong2023raft}, and DPO \citep{rafailov2023direct} in this work, to study the catastrophic forgetting issue of LLMs after RLHF.

\paragraph{Pretraining, fine-tuning, and distributional shift.}
Before the emergence of foundation models, the pre-training and fine-tuning paradigm had already achieved remarkable accomplishments across numerous applications \citep{he2016deep, radford2021learning, devlin2018bert}.  However, when deploying pre-trained models into real-world applications and fine-tuning them, a common challenge arises: encountering novel samples from a target distribution that differs from the fine-tuning distribution \citep{andreassen2021evolution, goyal2022finetune, zhang2022contrastive, lin2022bayesian, zhou2022model, zhou2022sparse, lin2022zin, tan2023provably}.
To address this issue, several approaches have been proposed.
For instance, \citep{Wortsman2021wiseft, cha2021swad, chu2022dna} suggest leveraging the weight ensemble of the pre-trained model and the fine-tuned model to enhance out-of-distribution (OOD) performance.
Another strategy, as proposed in \citep{kumar2022fine}, is the LP-FT technique, which involves initializing the pre-trained feature extractor with a reasonably good classifier.
This initialization is particularly important when the classifier is randomly initialized, as the pre-trained features can easily be distorted to accommodate the random classifier during fine-tuning, exacerbating the issue of catastrophic forgetting.

\paragraph{Catastrophic forgetting and continual learning.} DNN tends to lose the knowledge of previously learned task (e.g., pretraining task) when it begins to learn a new task (e.g., the fine-tuning task) \citep{mcclelland1995there}. Various attempts have been made to alleviate catastrophic forgetting. \citep{xuhong2018explicit, ritter2018online, aljundi2018memory,schwarz2018progress} impose a penalty on the change of the parameter on the new task. \citep{lifelong2021} transfers knowledge from related new knowledge types back to the old types by continually training the representations of old knowledge with the data for new knowledge using a self-training loss. \citep{SelfInfo2023} observes that LLMs tend to rely on pre-existing knowledge, neglecting recent facts and leading to incorrect reasoning chains that ultimately diminish the efficacy of
information updates, and proposes to mitigate exposure bias by incorporating the selection of relevant facts into training losses.
\citep{kirkpatrick2017overcoming} gain intuition from Taylor expansion of the losses of the old task at the point of fine-tuned parameter, and further proposes EWC by incorporating the Hassien matrix into parameter regularization.
The reply-based method tries to approximate and recover the old data distribution.
Popular methods in this direction include sampling methods which store a few old training samples with a small memory buffer \citep{vitter1985random, riemer2018learning, chaudhry2018efficient, cha2021co2l, caccia2021new}, and generative methods which generate samples from the old distributions with a generative model \citep{caccia2020online}. Knowledge distillation (KD) methods try to keep the prediction of the fine-tuned model close to that of the old model.
KD can be naturally combined with experience reply.
For example, \citep{rebuffi2017icarl} proposes to perform KD on the samples of new tasks as well as the old samples stored in the buffer.

Notably, previous continual learning focuses on sequentially learning tasks which learns a sequence of tasks in order and measures the forgetting of older tasks when learning new tasks \citep{wang2023comprehensive}.
Whereas, we focus on the generality forgetting of the pre-trained foundation model during fine-tuning a specific task.
\paragraph{Alignment tax.} \cite{ouyang2022training} reports that they observe significant alignment tax when developing InstructGPT. They have also tried to adopt Experience Replay to alleviate this issue, which is followed by \cite{zheng2023secrets}. However, we show  in Appendix~\ref{app:experience_replay} that Experience Relay is less favorable when compared with model averaging. \cite{noukhovitch2024language} tried to use stochastic weight averaging, which still under-performs our method as shown in Figure~\ref{fig:no_data_baseline}. \cite{li2024multi} finds that DPO induces less alignment tax compared with other RLHF algorithms, which is consistent with our findings (e.g., Figure~\ref{fig:adaptive_model_averaging}). \cite{askell2021general} reports that they didn't observe significant alignment tax when prompting LLM to align with humans. However, we focus on a more standard setting that the LLM is fully fine-tuned for RLHF.

\section{RLHF Basics}

Following \citep{ouyang2022training, bai2022training, dong2023raft, touvron2023llama, rafailov2023direct}, we assume that there exists a ground-truth reward function $r^*(x,a): \cX\times\cA \to [0,1]$ where $\cX$ and $\cA$ are the spaces of prompt and action. The preference ranking satisfies the Bradley-Terry model \citep{bradley1952rank}: the probability of $a^1 \in \cA$ being preferred is
\begin{equation}  \label{eqn:bt}
    \begin{aligned}
        \mathbb{P}(a^1 \succ a^2|x,a^1,a^2) =
        \frac{\exp(r^*(x,a^1))}{\exp(r^*(x,a^1)) + \exp(r^*(x,a^2))}.
    \end{aligned}
\end{equation}
We denote an LLM by a policy $\pi$ that maps $x$ to a distribution over the response space $\cA$. The main goal of RLHF is to align the staring checkpoint $\pi_{\thetasft}$ with the human preference so that it achieves high reward measured by $r^*$, but we may also impose additional constraints to avoid overfitting like requiring the models to stay close to the $\pi_{\thetasft}$. In practice, we learn from  a preference dataset of the form $\cD = \{(x, a_w, a_l)\}$, where $a_w$ is the preferred response. Typically, we will first train a reward model $r$ as the Maximum Likelihood Estimation \citep{ouyang2022training, bai2022training, touvron2023llama} on the preference dataset $\cD$ and then perform reward optimization by different algorithms.

\textbf{Rejection Sampling Finetuning} (RSF) is proposed in \citep{dong2023raft, touvron2023llama, yuan2023rrhf, gulcehre2023reinforced} with several variants. Essentially, the RSF learns from the best-of-n policy \citep{nakano2021webgpt}, which samples $n$ responses for each prompt query and returns the one with the highest reward. As suggested by \citep{dong2023raft, touvron2023llama, gulcehre2023reinforced}, we adopt an iterative training set-up for the implementation instead of always sampling the samples from the starting checkpoint because we find that the iterative training is far more sample-efficient. Specifically, for each iteration, we first sample a batch of prompts and generate $n$ responses for each prompt from current model. Then, we use the reward model to compute the rewards for each prompt-response pair and for each prompt, we select the one with the highest reward into a small subset. By this process, we collect a batch of samples from the best-of-n policy that are with high reward. We simply fine-tune the current model on this subset to get the next model and the next iteration begins.

\textbf{PPO} is the the classical method for RLHF and has gained its success in aligning Chat-GPT \citep{OpenAI2023GPT4TR}. In contrast to the implementation in traditional DRL scenario, for alignment of LLMs, following \citep{ziegler2019fine, wu2021recursively, ouyang2022training, rafailov2023direct, liu2023statistical}, we modify the reward optimization as the following KL-regularized form:
$$
    \tilde{r}(x,a) = r(x,a) - \eta \log \frac{\pi(a|x)}{\pi_{\thetasft}(a|x)},
$$
where $\eta > 0$ is a hyper-parameter to control the level of KL penalty.

\textbf{Direct Preference Optimization} (DPO) is proposed by \citep{rafailov2023direct} from the following KL-constraint optimization problem:
\begin{equation}\label{eqn:kl_constriant}
    \max_{\pi} \mathbb{E}_{x} 
    \mathbb{E}_{a \sim \pi(\cdot|x)}\left[ r^*(x,a) + \eta \log \frac{\pi_{\thetasft}(a|x)}{\pi(a|x)}\right].
\end{equation}
It is known that \eqref{eqn:kl_constriant} admits the following closed-form solution $\pi^*(\cdot|x) = \frac{1}{Z(x)}\pi_0(\cdot|x) \cdot \exp\Big(\frac{1}{\eta} r^*(x,\cdot)\Big)$ (see e.g. Proposition 7.16 of \citep{zhang_2023_ltbook}), where $Z(x)$ is the normalization constant. We can now represent $r^*$ by $\pi^*$ as follows:
$$
    r^*(x,a) = \eta \log \frac{\pi^*(a|x)}{\pi_0(a|x)} + \eta \log Z(x).
$$
Plugging the reparameterization of $r^*$ into the preference model in \eqref{eqn:bt}, we get
\begin{equation}\label{eqn:repara_bt_dpo}
    \mathbb{P}(a^1 \succ a^2 |x, a^1, a^2) = \frac{1}{1 + \exp \Big(\eta \log \frac{\pi^*(a^2|x)}{\pi_0(a^2|x)} - \eta \log \frac{\pi^*(a^1|x)}{\pi_0(a^1|x)}\Big)}.
\end{equation}
The idea of DPO is to find a model $\pi$ so that it maximizes the likelihood given in \eqref{eqn:repara_bt_dpo} on the offline preference dataset. Therefore, it chooses to minimize the following loss function:
\begin{equation} \label{eqn:dpo_loss}
    \begin{aligned}
        \mathcal{L}(\theta, \pi_{\thetasft}, \cD)= - \sum_{(x,a_w,a_l) \in \cD} \Big[ \log \sigma\Big(\eta \log \frac{\pi_{\theta}(a_w|x)}{\pi_{\thetasft}(a_w|x)} - \eta \log \frac{\pi_{\theta}(a_l|x)}{\pi_{\thetasft}(a_l|x)} \Big)\Big],
    \end{aligned}
\end{equation}
where the reward modeling step is bypassed.
\subsection{Algorithm of Heterogeneous Model Averaging}

\textbf{Reward Preserving Updating.}~It is noteworthy that Eqn.~\eqref{eqn:adaptive_model_averaging} represents a RL problem. To implement Eqn.~\eqref{eqn:adaptive_model_averaging}, RL algorithms such as RSF, PPO, or DPO need to be implemented, involving extra implementation details that depend on the algorithm. To address this issue, we propose a proxy distillation method. Specifically, given a policy $\pi_{\theta}$ after RLHF, we generate a proxy dataset by
\begin{align}
    \label{eqn:d_theta}
    \cD_\theta = \{ (x, a): a \sim \pi_{\theta} (\cdot | x), \mbox{ for } x \in \cX \}.
\end{align}
Since the data in $\cD_\theta$ is generated by $\pi_{\theta}$, this data should have a high reward. Therefore, maximizing the likelihood on $\cD_\theta$ could result in a model with a high reward. Specifically, we optimize the following
\begin{align}
    \label{eqn:adaptive_model_averaging_distll}
    \max_{\alpha_1, ..., \alpha_K \in \Omega} \frac{1}{|\cD_{\theta}|} \sum_{(x, a) \in \cD_{\theta}} \log [\pi_{\theta(K)}(a|x)].
\end{align}

The algorithm of Heterogeneous Model Averaging is summarized as follows:
\vspace{-.8em}
\begin{algorithm}[H]
    \caption{HMA: Heterogeneous Model Averaging}
    \label{algo:hma}
    \begin{algorithmic}[1]
        \label{algo:HMA}
        \renewcommand{\algorithmicrequire}{\textbf{Input:}}
        \renewcommand{\algorithmicensure}{\textbf{Output:}}
        \REQUIRE The reward model $r(\cdot, \cdot)$, initial policy $\pi_{\thetasft}$, prompt set $\cD_x$, hyper-parameter $K$, merge ratio $\alpha$.
        \ENSURE  The output policy $\pi_{\theta(K)}$.
        \\ \STATE Perform vanilla RLHF by Eqn~\eqref{eqn:RLHF_basic} and obtain $\pi_{\theta}$.
        \\ \STATE Distill $\cD_\theta$ from $\pi_{\theta}$ according to Eqn.~\eqref{eqn:d_theta}.
        \\ \STATE Initialize $\alpha_1, ..., \alpha_K \in [0, 1]$ for the $K$ parts of the Transformer, respectively.
        \\ \STATE Obtain the averaged model $\theta(K)$ with $\alpha_1, ..., \alpha_K $.
        \\ \STATE Solve Heterogeneous ratios $\alpha_1, ..., \alpha_K$ according to Eqn.~\eqref{eqn:adaptive_model_averaging_distll}.
        \\ \STATE Return the $\theta(K)$ with the optimized $\alpha_1, ..., \alpha_K$.

    \end{algorithmic}
\end{algorithm}
\vspace{-1em}

\section{More Results}
\subsection{Experience Replay}
\label{app:experience_replay}
 In our alignment tax situation, we aim to preserve a wide range of abilities gained during pre-training. It is possible to  replay a small subset of pretraining data, which also known as Experience Replay (ER) \citep{rebuffiincremental,shin2017continual}. However, this method is  less practical since pre-training datasets of most models are often not publicly available. Further more, even if we can access the pre-training data, retaining a subset of the pre-training data entails extra computational costs and implementation intricacies, making it less preferable \citep{noukhovitch2023language}. In this part, we compare ER with MA. Specifically, we include a small proportion of randomly subsampled pre-training data during the RLHF stage. Here, we denote $\cD_{{pre}}$ as the pre-training data distribution, and our objective is to solve the following:
\begin{align*}
    \max_{\theta} \mathbb{E}_{x} \mathbb{E}_{a \sim \pi_\theta(\cdot|x)} [ r^*(x,a)] + \lambda \bbE_{(x, a) \sim \cD_{{pre}}} \log \pi_{\theta}(a|x)
\end{align*}
We experiment with different penalty weights $\lambda$ such as $0.25$, $0.5$, $1$, $2$, and $4$. Importantly, we utilize the data proportion as a proxy for setting the penalty weight. For instance, we do not explicitly apply a penalty of 4 when $\lambda=4$; instead, we include 4 times the replay data over the RLHF data in a batch. Refer to the Appendix~\ref{app:impl_details} for more details.

\begin{figure*}
    \centering
    \includegraphics[width=0.3\linewidth]{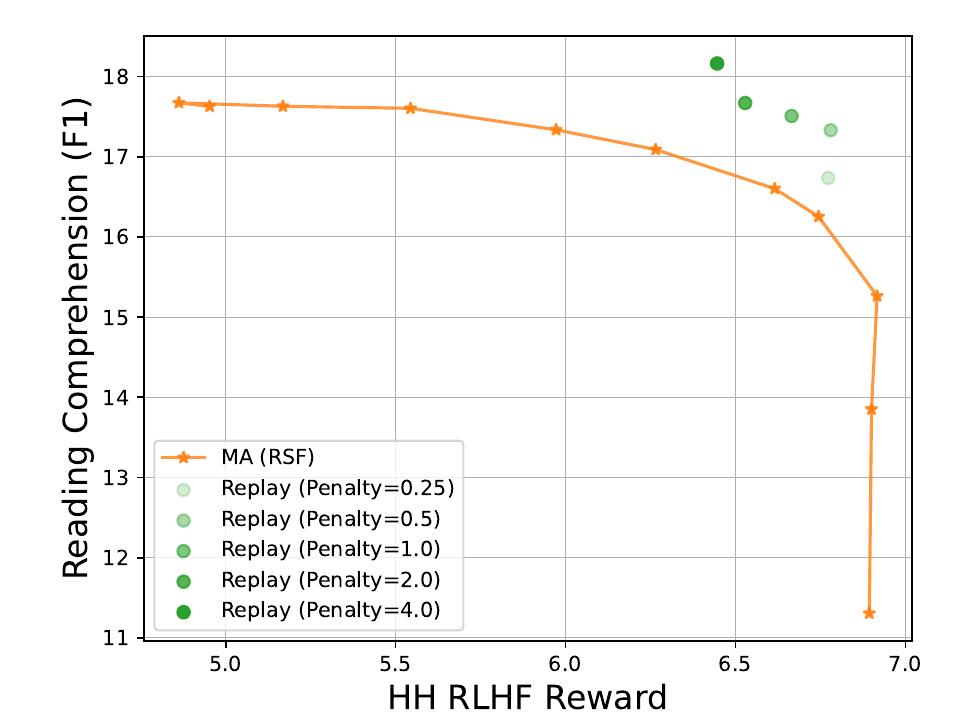}
    \includegraphics[width=0.3\linewidth]{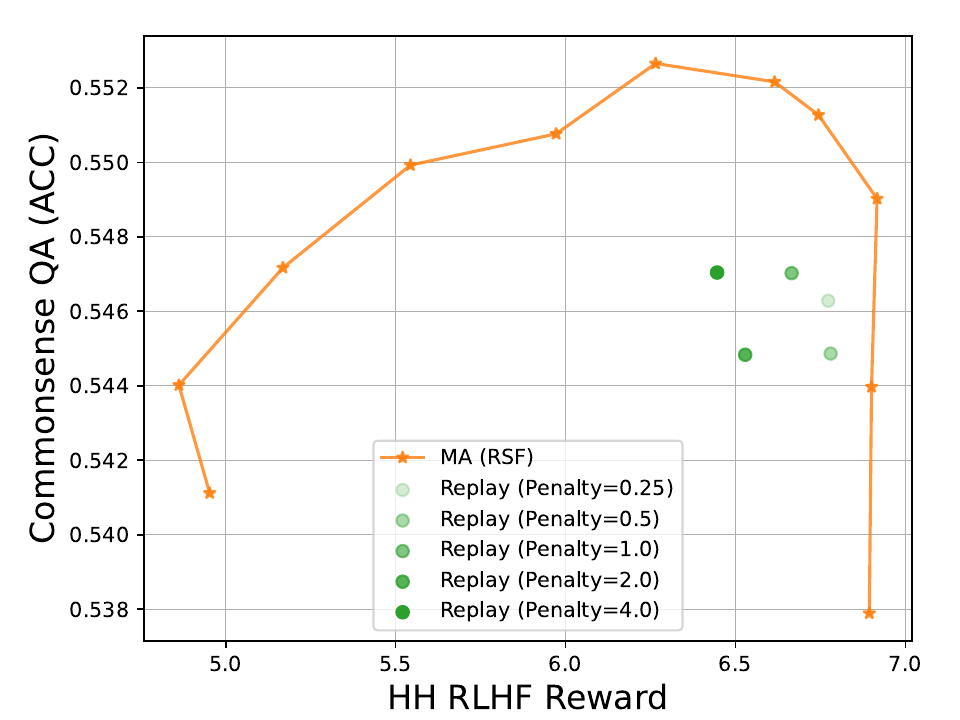}
    \includegraphics[width=0.3\linewidth]{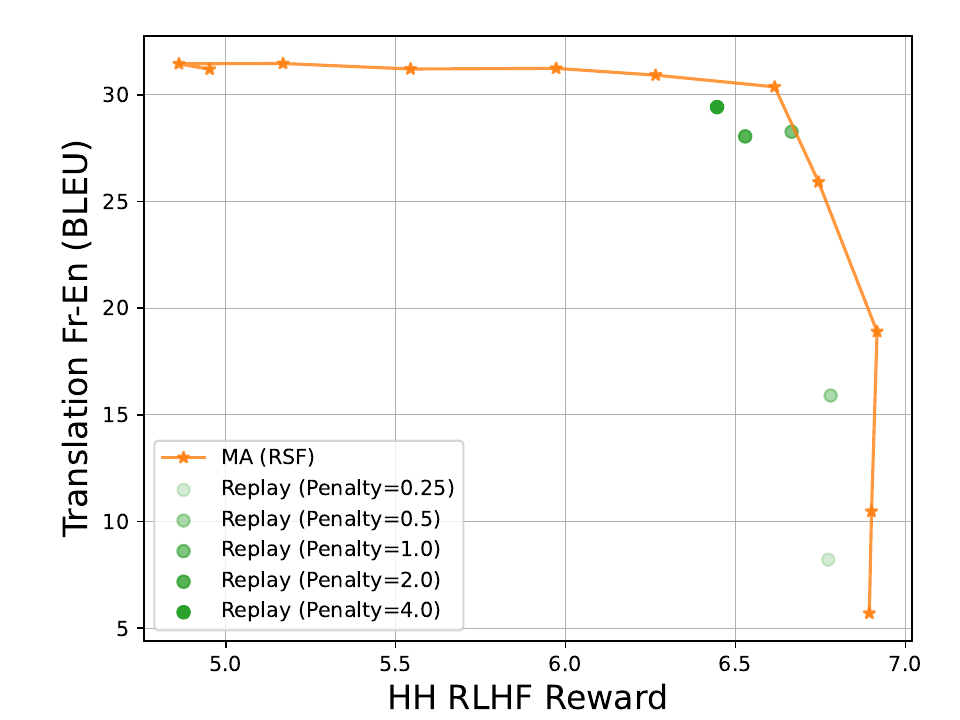}
    \vspace{-1em}
    \caption{
    Comparison of model averaging with Experience Replay.}
    \label{fig:relay_baseline}
    \vspace{-1em}
\end{figure*}

\textbf{Results.}~The results of ER are displayed in Figure~\ref{fig:relay_baseline}. Additionally, we include the performance of model averaging for comparison. It is evident that while ER has access to pre-training data, it only demonstrates superior performance over model averaging in the Reading Comprehension dataset (Figure~\ref{fig:relay_baseline} - Left), and falls short of model averaging in the Commonsense QA (Figure~\ref{fig:relay_baseline} - Middle) and Translation (Figure~\ref{fig:relay_baseline} - Right) benchmarks.

\textbf{Discussion of ER results.}~The differing performance of ER compared to model averaging is somewhat surprising. Despite maintaining extra pre-training data, which is four times larger than the RLHF data (400M token), ER under-performs model averaging in two out of three benchmarks. This may be attributed to the vast size of the pre-training data (1.2T token), such that even when replaying a subset four times larger than the RLHF data, it only covers about 0.03\% of the pre-training data.  Consequently, the data corresponding to certain abilities may be underrepresented in the replay dataset. With a substantial pre-training dataset and a wide range of abilities to preserve, it becomes challenging to maintain all abilities through replay.

\vspace{-.5em}
\subsection{Reward Penalty}
\label{app:reward_penalty}
\begin{figure}
    \centering

    \includegraphics[width=0.6\linewidth]{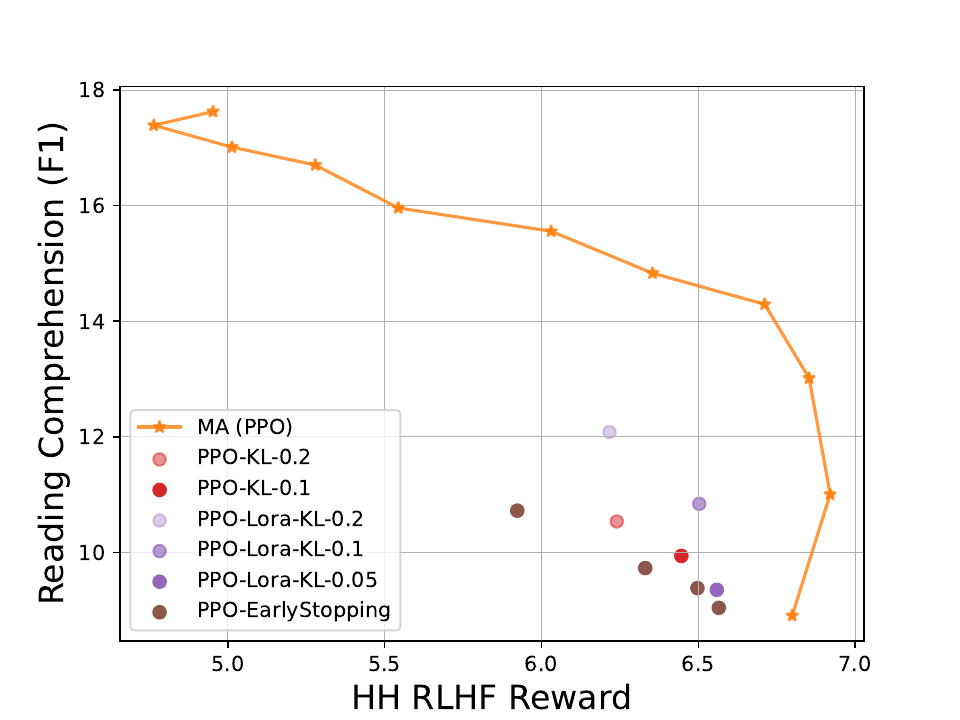}
    \caption{Comparison of model averaging with reward penalty for PPO.}
    \label{fig:PPO_reward_penalty_results}
\end{figure}

It is a common practice to impose Kullback–Leibler (KL) penalty on the RL reward in the PPO. Such a penalty can also regularize the policy to stay closer to the initial policy, which in return can reduce the alignment tax. Following \citep{ziegler2019fine, wu2021recursively,ouyang2022training,yuan2023rrhf}, we modify the raw reward function with an additional KL penalty \citep{ziegler2019fine}.
\begin{align} \label{eqn:kl_constriant_method}
    \max_{\pi} \mathbb{E}_{x} \mathbb{E}_{a \sim \pi_\theta(\cdot|x)} [r^*(x,a)] - \mathrm{KL} (\pi_\theta||\pi_{\thetasft}) ,
\end{align}
where we use $\mathrm{KL} (\pi_\theta||\pi_{\thetasft})$ to denote $\bbE_{x} [\mathrm{KL} (\pi_\theta(\cdot|x)||\pi_{\thetasft} (\cdot|x) )]$ for short.  We compare vanilla model averaging methods with the reward penalty by considering different KL penalties in $\{0.05, 0.1, 0.2\}$. The results are shown in Figure~\ref{fig:PPO_reward_penalty_results}. We can see that while a larger KL penalty can partially mitigate the forgetting issue, the model averaging is much more effective than the reward penalty in terms of the alignment-forgetting trade-off.

\subsection{Consistency of different combination ratios among various tasks}
\label{app:proof_of_concept_combination}
\begin{figure}[H]
    \centering
    \includegraphics[width=0.3\linewidth]{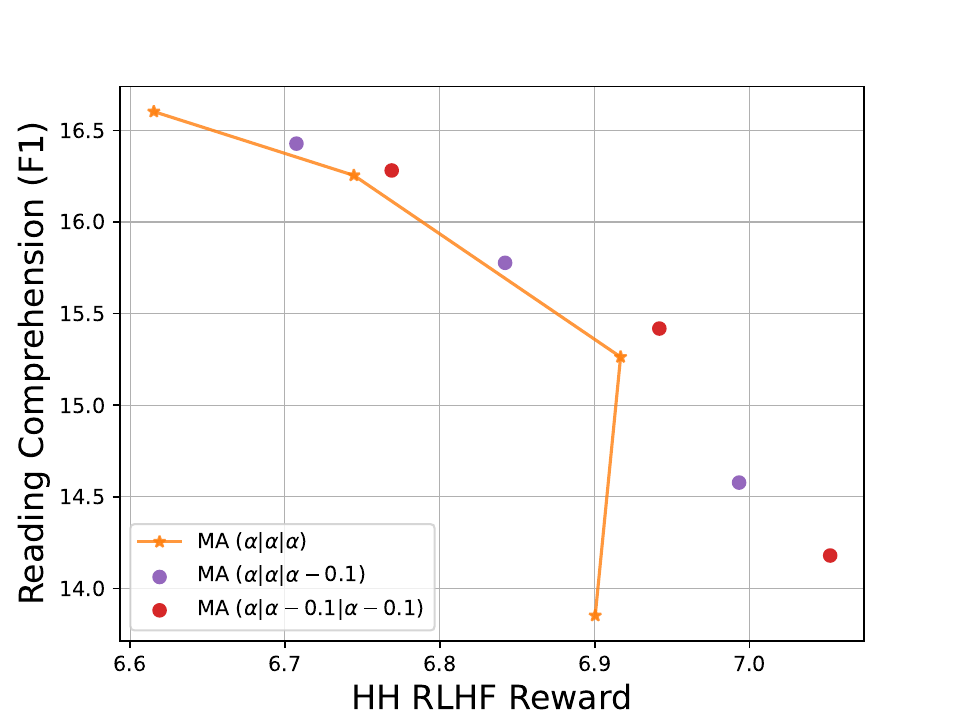}
    \includegraphics[width=0.3\linewidth]{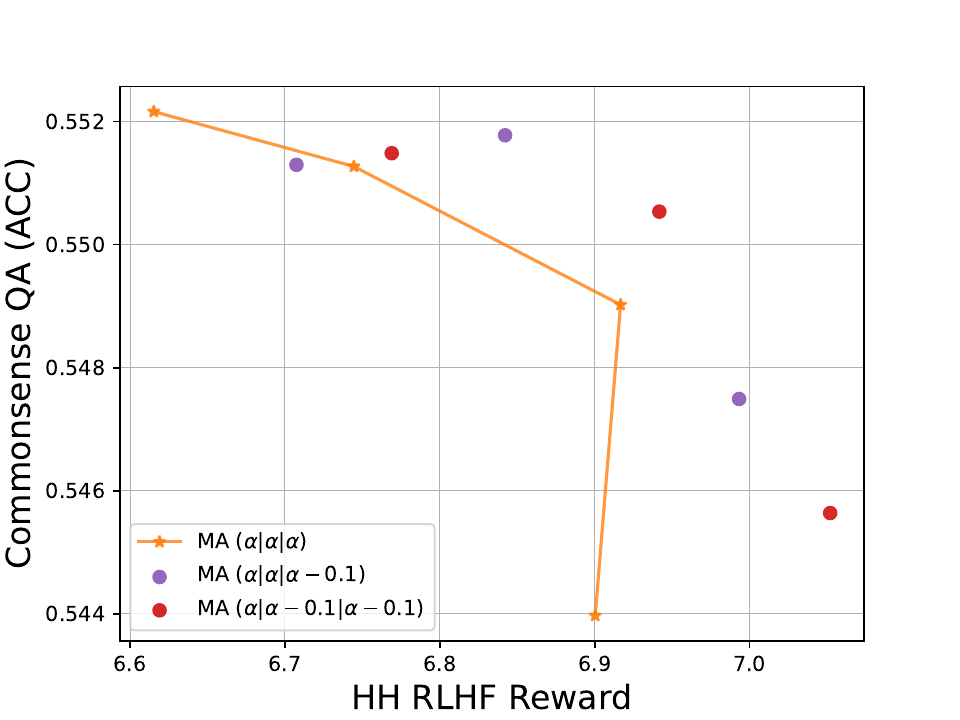}
    \includegraphics[width=0.3\linewidth]{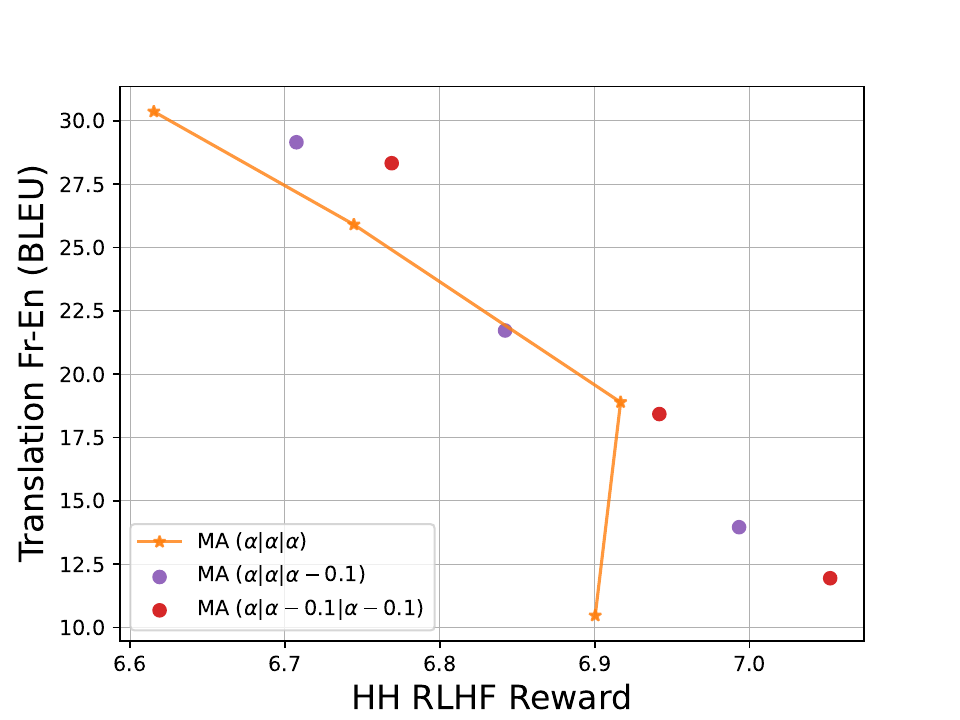}
    \vspace{-1em}
    \caption{Evaluation of different combination ratios.}
    \label{fig:proof_of_concept}
    \vspace{-1em}
\end{figure}

 We try three patterns experiment given a base $\alpha \in \{0.2, 0.3, 0.4\}$ :  (a) $\alpha_1 = \alpha_2 = \alpha_3 = \alpha$; (b) $\alpha_1 = \alpha_2 = \alpha$, $\alpha_3 = \alpha - 0.1$, and (c) $\alpha_1 = \alpha$, $\alpha_2 = \alpha_3 = \alpha - 0.1$. We use $(\alpha |  \alpha | \alpha)$, $(\alpha |  \alpha | \alpha-0.1)$ and $(\alpha | \alpha-0.1| \alpha-0.1)$  to denote these three patterns, respectively. These results confirm that certain ratio combinations exceed the trade-off curve of vanilla model averaging, as displayed in Figure~\ref{fig:proof_of_concept}.
Notably, some combination ratios consistently outperform the equal ratio across various benchmarks. This affirms the potential to identify consistent combination ratios that demonstrate superior performance across a broad spectrum of benchmarks in terms of alignment-forgetting trade-off.

\subsection{Results of $\alpha=0.2$}
\label{app:alpha_02}
The following results show that when we chose $\alpha=0.2$, MA and HMA consistently alleviate the alignment tax without sacrificing any alignment performance. 
\begin{figure}[H]
    \centering
    \includegraphics[width=0.3\linewidth]{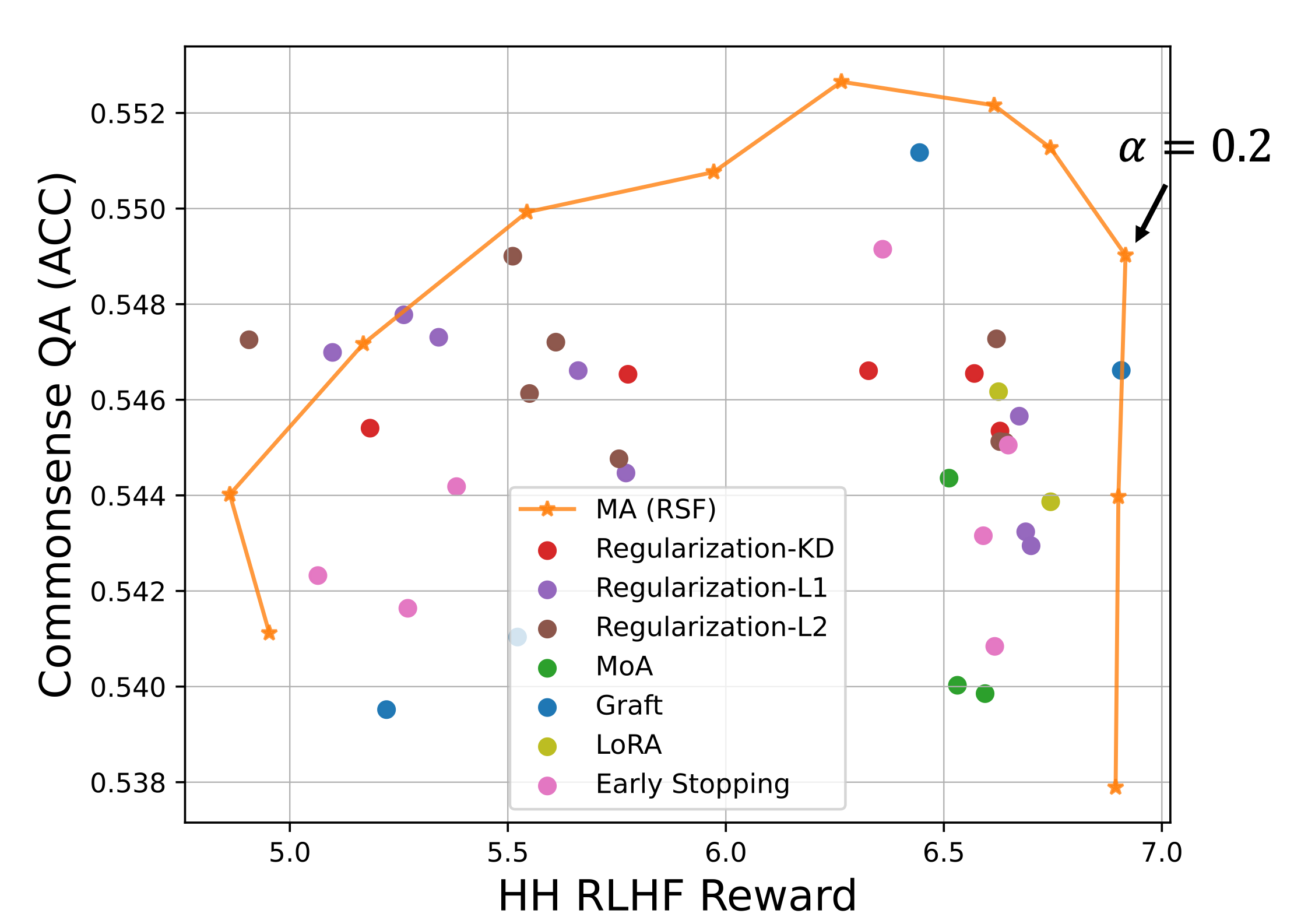}
    \includegraphics[width=0.3\linewidth]{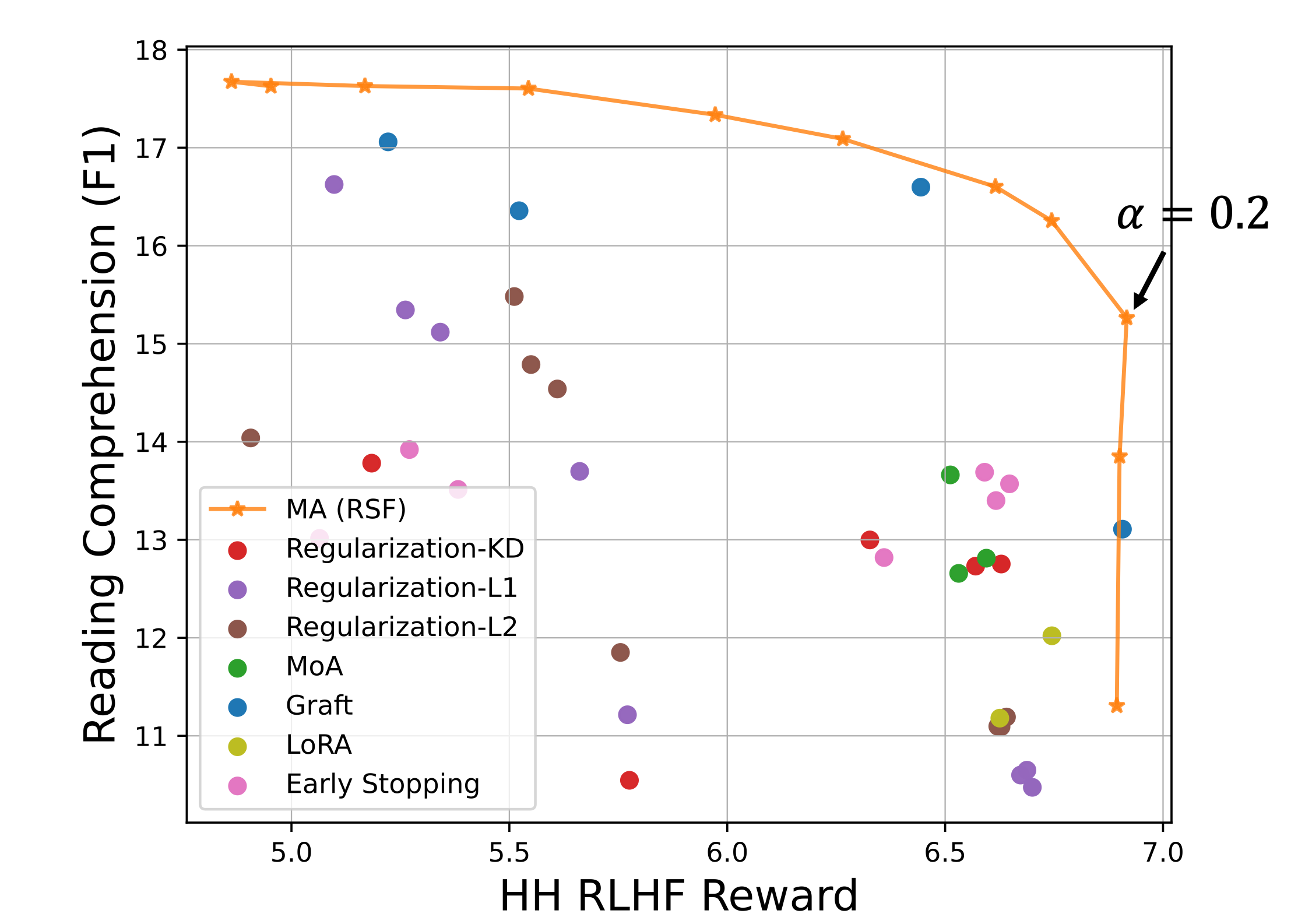}
    \includegraphics[width=0.3\linewidth]{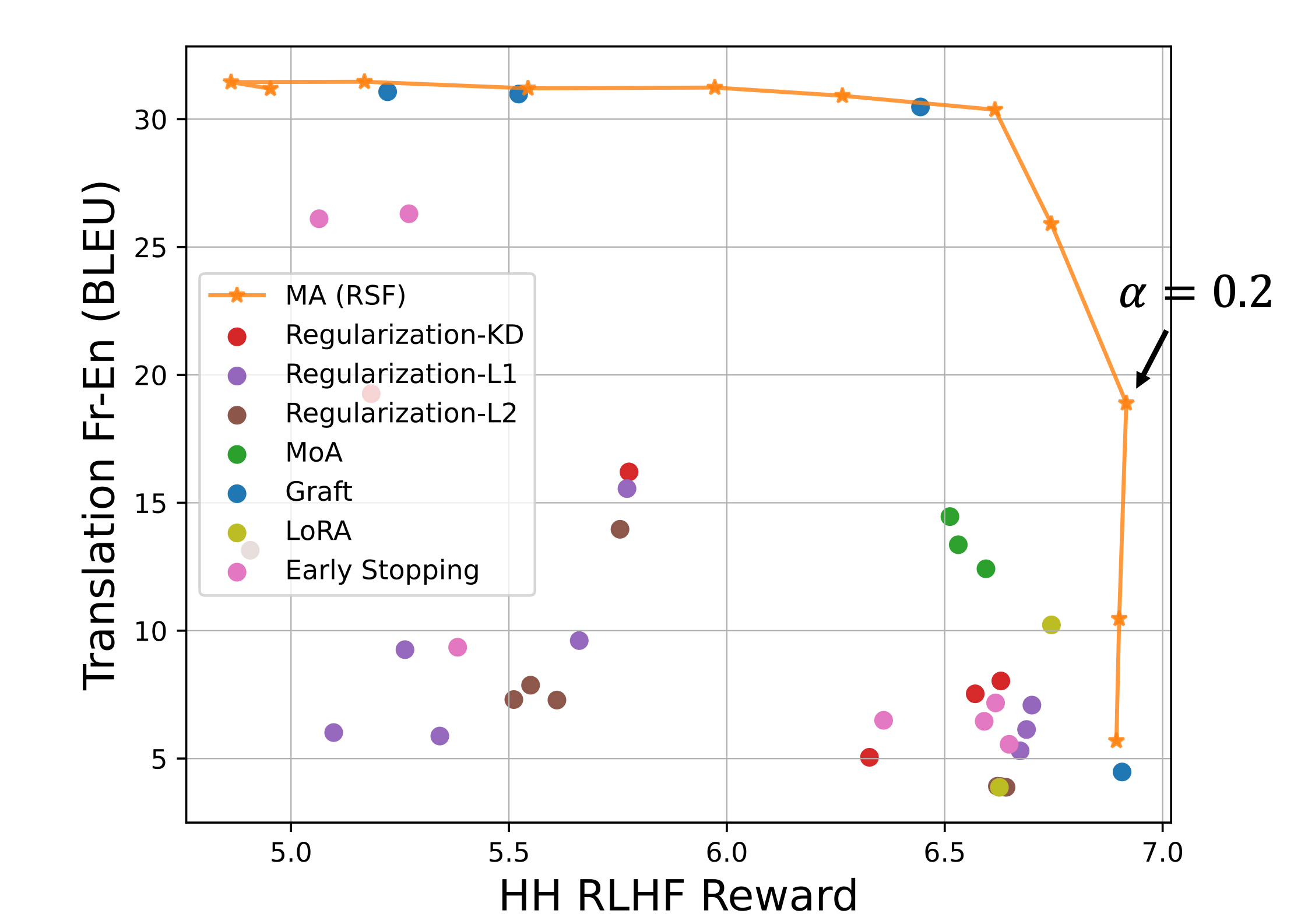}
    \caption{Illustration of $\alpha=0.2$ on vanilla model averaging}
    \label{fig:alpha-0.2_1}
\end{figure}
\begin{figure}[H]
    \centering
    \includegraphics[width=0.3\linewidth]{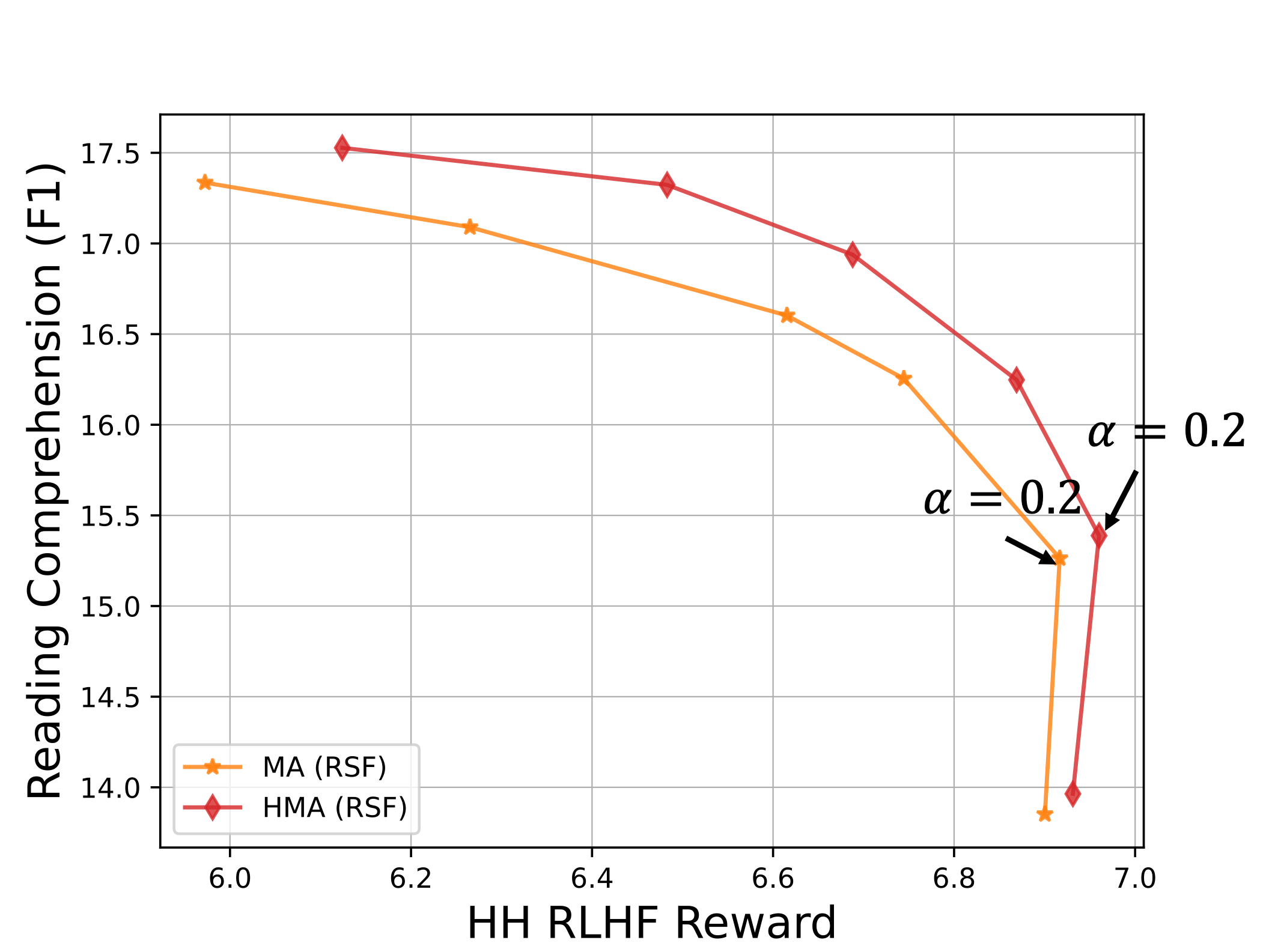}
    \includegraphics[width=0.3\linewidth]{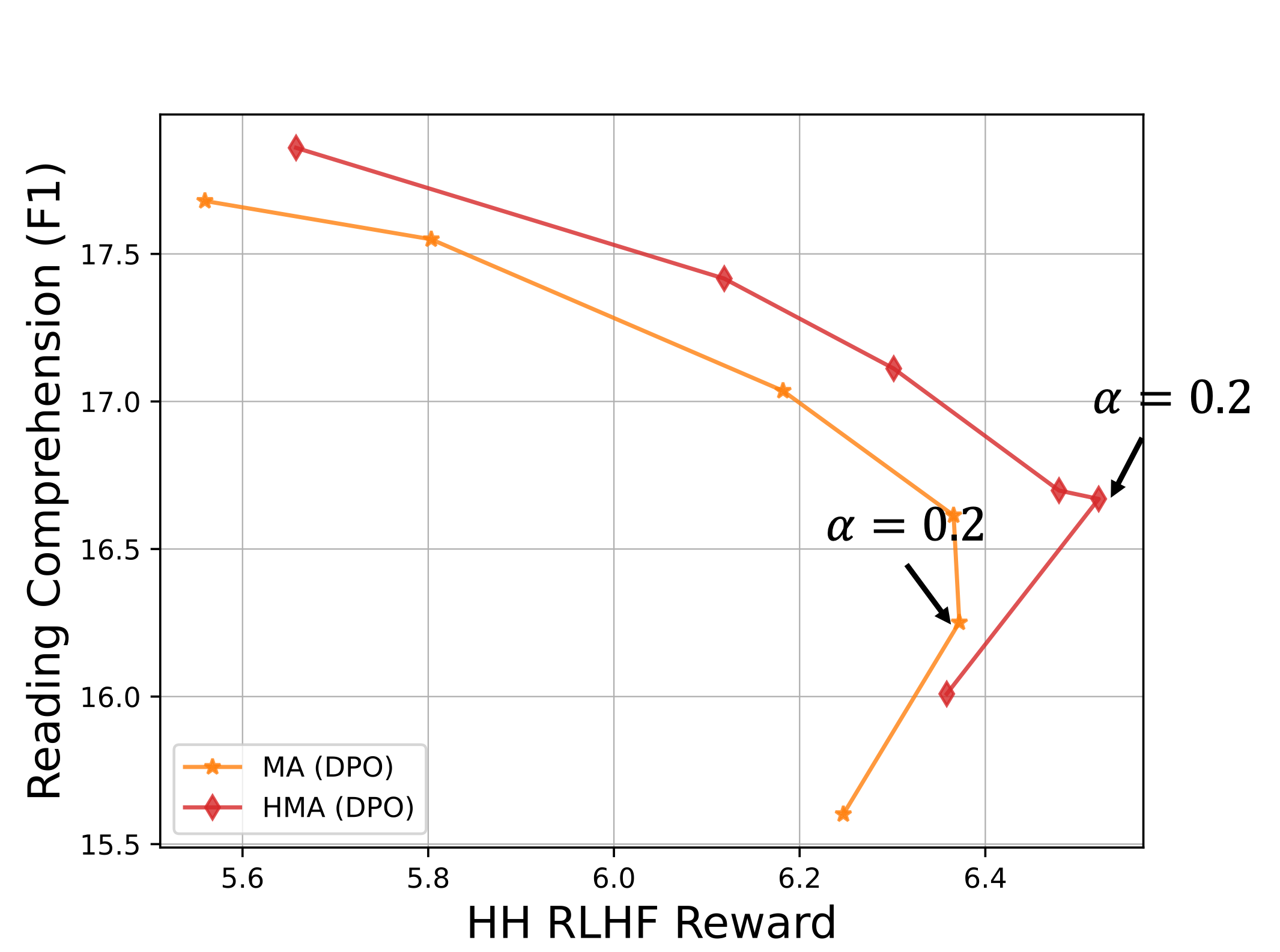}
    \caption{Illustration of $\alpha=0.2$ on HMA}
    \label{fig:alpha-0.2_2}
\end{figure}

\section{Implementation Details}
\label{app:impl_details}

In this section, we introduce the implementation details for the methods mentioned in Section~\ref{sect:basic_settings}.

\subsection{Rejection Sampling Fine-tuning Implementation}
\label{app:rsl_details}

The rejection sampling fine-tuning (RSF) is proposed in \citep{dong2023raft, touvron2023llama, yuan2023rrhf, gulcehre2023reinforced} with several variants. Essentially, RSF earns from the best-of-n policy \citep{nakano2021webgpt}, which samples $n$ responses for each prompt query and returns the one with the highest reward. In this work, we implement the algorithm with the official code provided in LMFlow\footnote{\url{https://github.com/OptimalScale/LMFlow}}. We adopt most of the hyper-parameters as suggested by \citep{dong2023raft} and focusing on tuning the learning rate by searching over $\{1 \times 10^{-6}, 2\times 10^{-6}, 1 \times 10^{-5}\}$ and $1 \times 10^{-5}$ is taken for our main experiments.

As suggested by \citep{dong2023raft, touvron2023llama, gulcehre2023reinforced}, we adopt an iterative training set-up for the implementation instead of always sampling the samples from the starting checkpoint because we find that the iterative training is far more sample-efficient. Specifically, for each iteration, we first sample a batch ($2048$) of prompts and generate $n=32$ responses for each prompt from current model. Then, we use the reward model to compute the rewards for each prompt-response pair, and for each prompt, we select the one with the highest reward into a small subset. Through this process, we collect $2048$ samples from the best-of-32 policy that are with high reward. We simply fine-tune the current model on this subset to get the next model and the next iteration begins.

When RSF is combined with other methods for preventing the model from forgetting, we follow \citep{touvron2023llama, dong2023raft} to align the models in a distillation style. Specifically, we run RSF algorithm as described above until the model converges to a rather stable level of reward. Then, we collect the best-of-32 samples along the way of training and fine-tune the model from the starting checkpoint with the additional methods for mitigating the forgetting issue. In comparison, we note that \citep{touvron2023llama} only uses the largest 70B Llama 2-Chat models to collect best-of-n samples and other smaller models are then fine-tuned on these collected data and \citep{dong2023raft} uses LLaMA-7B to run RSF and uses the collected data to fine-tune other LLMs.

\subsection{Implementation of PPO}

The experiments with PPO in this work are conducted using the open-source package Transformer Reinforcement Learning (TRL)\footnote{\url{https://github.com/huggingface/trl}}. It is known that the PPO is significantly less stable as compared to supervised learning \citep{choshen2019weaknesses} and sensitive to the hyper-parameter and code-level optimization \citep{engstrom2020implementation}. To tune PPO to its best performance, we include several empirical enhancements and we record our tuning process, as well as the successful/unsuccessful attempts in this subsection for interested readers.

First, we follow \citep{ramamurthy2022reinforcement} to warm up by finetuning the model on the preferred samples of the preference dataset for 1 epoch for a more stable training process. Moreover, in contrast to the implementation in traditional DRL scenario, for alignment of LLMs, following \citep{ziegler2019fine, wu2021recursively, ouyang2022training, rafailov2023direct, liu2023statistical}, we will also modify the reward optimization as the following KL-regularized form:
$$
    \tilde{r}(x,a) = r(x,a) - \eta \log \frac{\pi(a|x)}{\pi_0(a|x)},
$$
where $\eta > 0$ is a hyper-parameter to control the level of KL penalty.

However, even though we first finetune the models with the preferred samples and train with an additional KL penalty, the PPO training can still lead to an unstable reward level and failure. For the first issue, with the ultimate hyper-parameter, we will run PPO with three independent seeds and take the best models. We now focus on the second issue. One notable failure signal of PPO training is that the models suddenly refuse to answer the question (prompt), or reply with incomplete sentences, which may be detected by (1) a shorter average response length; (2) the incomplete sentences in randomly displayed sample responses within one iteration; (3) sudden drop in reward value. Once such a drop happens, the models just collapse and the training fails.

\textbf{Hyper-parameter tuning.} To mitigate this issue, we carefully tune the learning rate, KL coefficient, update epoch, batchsize by grid search. We observe that for full training (without LoRA), a learning rate with $1 \times 10^{-6}$ is most suitable in terms of the trade-off between reward learning and training stability. Update epoch $=2$ performs best in our preliminary experiments for parameter tuning. A batchsize that is too large ($2048$) or too small ($128$) leads to unstable training. Therefore, we fix the batchsize as $512$ and the update epoch as $2$ to further tune the KL coefficient and learning rate. Ideally, in the mathematical formulation of KL-constrained RLHF, a smaller KL coefficient should lead to a higher reward value. In practice, we observe that for KL coefficient $\beta \in [0.05, 0.3]$, a smaller KL coefficient leads to a higher ultimate reward value of the obtained policy. However, for $\beta < 0.05$, the model collapses before it achieves the highest reward possible, leading to a even worse model compared to $\beta = 0.05$. The results are observed across more than 20 independent runs. Therefore, in the ablation study of the impact of KL coefficient for PPO, we choose $\beta = 0.05$ as the smallest KL coefficient. We mention in passing that due to the same instability issue, the LoRA training may also achieve better reward because we can optimize the model well with LoRA, while the full-trained models collapse before it achieve its best performance.

\textbf{Restart trick in critic training.} To further understand the reason why the PPO fails, we examine several training records provided by wandb. We found that before (or simultaneously) the models collapse, the critic loss increases significantly. After looking at the source code of TRL, we notice that there is a scaling factor of the critic loss of $0.1$, which may also suggest that the training processes of the critic and actor are different. Motivated by these observations, we try out different learning rates for the critic: (1) a larger learning rate for the critic; (2) a smaller learning rate for the critic; (3) decay/increase the learning rate of the critic every $10$ batch of the training. Unfortunately, we do not see significant improvement in either the training stability or the ultimate reward value. We noticed that the instability from value estimation (critic training) seems to be a well-studied problem in the DRL literature. For instance, \citep{lee2022offline} proposes to use a pessimistic (conservative) reward signal, which is obtained by reward model ensemble, which is also recommended in theoretical RLHF studies \citep{zhu2023principled, Xiong2023GibbsSF}. However, this requires to load multiple reward models at the same time, which is infeasible for us due to computational constraint. Motivated by the trick of PaLM (in the pre-trained stage) \citep{chowdhery2023palm}, which call back whenever the spikes happen in the loss curve, we simply train the model twice. Specifically, we run PPO training first and save the intermediate models for every iteration. Once the model collapses, we simply restart from a model $3$ iterations before the training fails and re-initiate the critic model. Then, we skip the actor training for $1$ iteration as a warm-up stage of the restarted critic. We observe that though the training still collapses easily after $10$-$20$ iterations of training, we do achieve a much higher reward value.

It is also interesting to design new algorithms to mitigate the value estimation error for a more stable PPO-based training, and we leave it for future study since it is beyond the scope of this work.

\subsection{Implementation of DPO}

We implement DPO by the open-source package Transformer Reinforcement Learning (TRL). We mainly use $0.1$ in our experiments but also try out $0.3$ and $0.5$ since the authors of original paper recommend to set it from $0.1$ to $0.5$. Then, we mainly tune the learning rate. We use the evaluation loss (which generally aligns with the evaluation accuracy) on the validation set of reward modeling for the model selection. We observe that for learning rate in $\{1\times 10^{-6}, 2\times 10^{-6}, 1\times 10^{-5}\}$, $1 \times 10^{-6}$ achieves the lowest evaluation loss so it is adopted in our experiments. We train DPO for up to $3$ epochs and evaluate the model every $0.5$ epoch by the evaluation loss on the validation set. The lowest evaluation loss and highest evaluation accuracy are achieved at the end of the first epoch so we use the model as the representative model of DPO though we do observe the validation reward of the model at $0.5$ epoch of the training is slightly higher. We suspect that this is because the equivalence of reward modeling and policy training are equivalent for DPO only when the optimization error is zero (see \citep{rafailov2023direct, azar2023general} for a detailed proof). In practice, since the samples are finite and we may not solve the non-convex optimization by finding its exact minimizer, the reward of the generator may not align with the accuracy of the discriminator (reward model).

\subsection{Implementations of Existing Methods to Alleviate Alignment Tax}
We test existing methods mainly on the RSF method which is implemented as discussed in Appendix~\ref{app:rsl_details}.
Details about how we implement existing methods to mitigate forgetting are described as follows.
\begin{itemize}
    \item [(a)] Early Stopping: The whole RSF is conducted for 10 iterations and we choose the model of RSF at numbers of iterations of 2,4,6,8 as the early stopping checkpoints.
    \item [(b)] Regularization towards $\thetasft$ in the weight space: For these kinds of methods. We alternative the training loss at the SFT stage in RSF by adding the regularization terms with different penalties. Specifically, we test $\{0.04, 0.1, 0.4, 0.6, 1\}$ for the L1 penalty and $\{0.01, 0.04, 0.06, 0.08, 0.1\}$ for L2 penalty.
    \item [(c)] Low-Rank Adaptation (LoRA): We implement two levels of LoRA. The typical version only considers the low-rank adaptation of MLP blocks and we have tested several ranks for 16-512, while only rank 512 gives a reasonable performance on the final alignment result. The other is the low-rank adaptation of both MLP and attention blocks, in this case, rank 16 makes a good performance on alignment.
    \item [(d)] Knowledge distillation: The implementation of this approach is similar to the Regularization method. We add the knowledge distillation term as a regularization term in the SFT stage. The penalty used here are $\{10^{-5}, 10^{-3}, 10^{-1}\}$.
    \item [(e)] Model Averaging: We simply interpolate the modules of linear layers in the whole model, e.g., Q, K, V projection layers in attention and MLP layers. We will vary the $\alpha$ from 0 to 1. The start point of the model averaging is the model after instruction following and the end point of that is the model after RLHF.
\end{itemize}

For the experience replay (ER) method, we uniformly sample the pre-trained data of Open-LLaMA-3B according to the penalty. Specifically, given the alignment data of 400M tokens and a penalty of 2, we will sample 800M token data from the pre-trained data. And then add data to conduct the SFT loss as a penalty.

\subsection{Implementations of Heterogeneous Model Averaging}

Notice that it is difficult to directly solve the Eqn.~\eqref{eqn:adaptive_model_averaging_distll} on the support set $\Omega$. So instead of directly optimizing the $\alpha_1, \dots, \alpha_K$, we reparameterize the $\alpha_1, \dots, \alpha_K$ as follows,
\begin{align}
    \hat{\alpha_i} = \sigma(s_i) + \epsilon; \quad \alpha_i = \frac{\hat{\alpha_i}}{\sum_{i=1,\dots,K}\hat{\alpha_i}} \alpha \label{eqn:ama_reparam}
\end{align}
where $\sigma(x) = \frac{1}{1+\exp(-x)}$ is the sigmoid function $s_i$ can take any real number. For each $s_1, \dots, s_K$, we can easily find the corresponding $\alpha_1, \dots, \alpha_K$ of Eqn.~\eqref{eqn:ama_reparam} belongs to the $\Omega$. In this way we can optimize on $s_1, \dots, s_K$ rather than $\alpha_1, \dots, \alpha_K$. Moreover, the $\epsilon$ in Eqn.~\eqref{eqn:ama_reparam} can serve as a boundary control parameter, that is, if we set $K=3, \epsilon=1$, then each $\alpha_i$ can just take values over $[0.2\alpha, 0.5\alpha]$. In practice, we will search the $\epsilon \in \{0, 0.1, \dots, 0.9\}$ to get the best model.

To get $D_\theta$, we will use the prompts from the training RLHF dataset to generate the full response with different policy $\pi_{\theta}$. Then we sample about 2000 pieces generated responses from the set consisting of the 5000 samples with the highest rewards. Then we can just take the $s_1, \dots, s_K$ as the optimization parameters and just finetuning them on the $D_{\theta}$.

Besides directly optimizing the Eqn.~\eqref{eqn:adaptive_model_averaging_distll}, we also test adding regularization terms of $\alpha_1, \dots, \alpha_K$. Generally we just add weighted L1 loss $\sum_i w_i|\alpha_i - \alpha|$ as the regularization terms. $w_i$ is chosen to make the middle part of the module change not too much.

Typically, we only average the weights in the linear layers and the $\alpha_1, \dots, \alpha_K$ works on transformer layers which contain self-attention and MLP. For the head layer, we just set the average weight as $\alpha$.

We give the hyper-parameters for the optimization in Table~\ref{tab:hyper_exp_ama}

\section{More Results}
\subsection{The Alignment Tax during Training (Results of Early Stopping)}
\label{app:alignmnet_tax_during_training}
The following figure shows the  RLHF reward and alignment tax during different training steps.

\begin{figure} [H]
    \centering
    \includegraphics[width=0.3\linewidth]{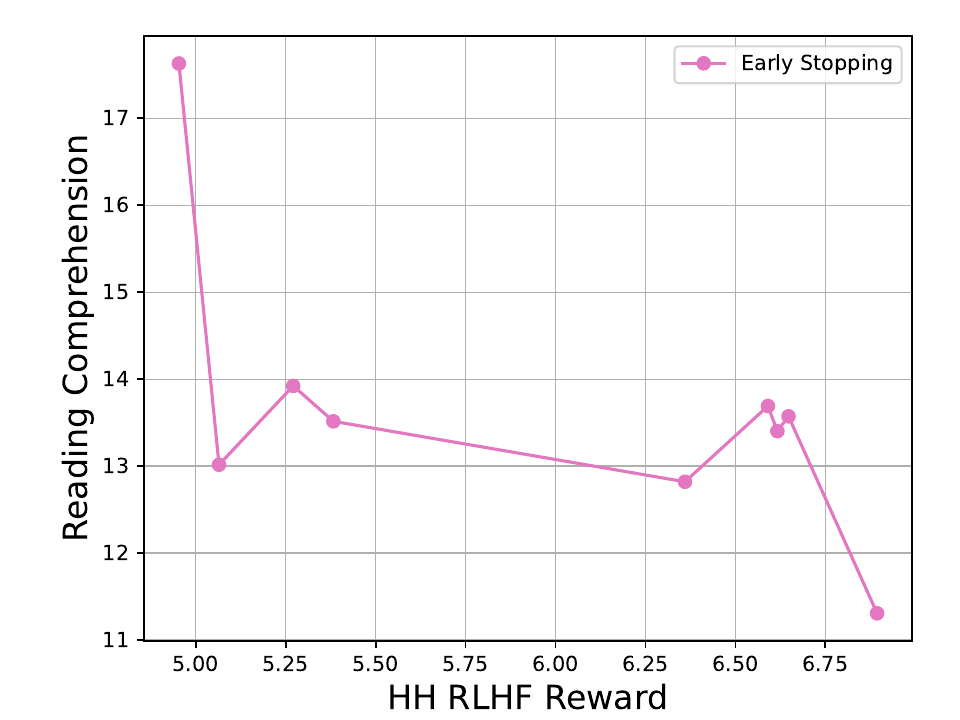}
    \includegraphics[width=0.3\linewidth]{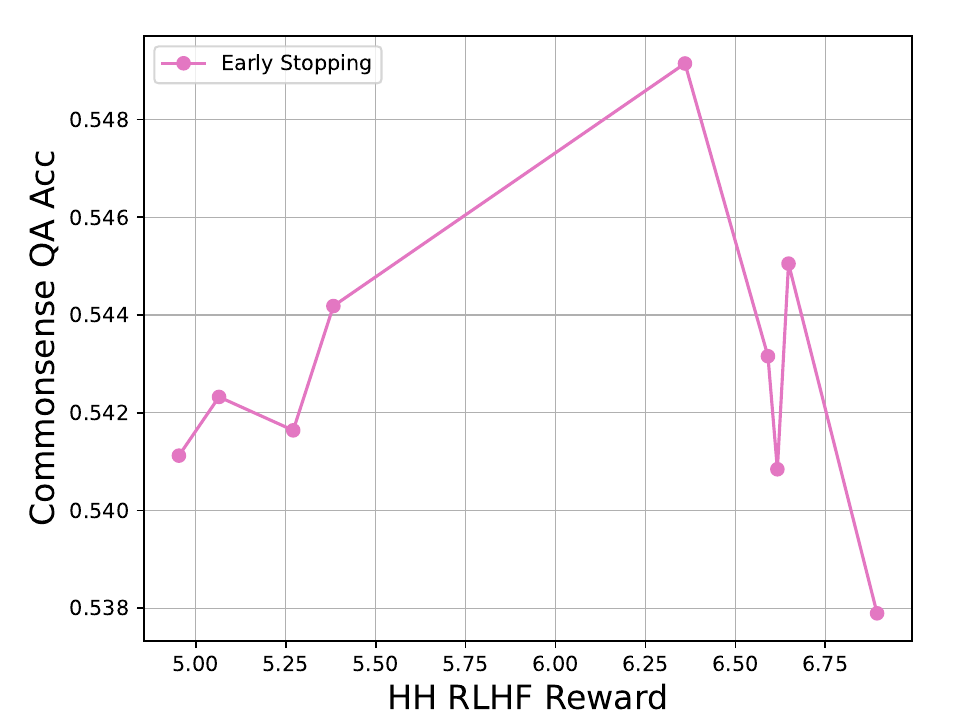}
    \includegraphics[width=0.3\linewidth]{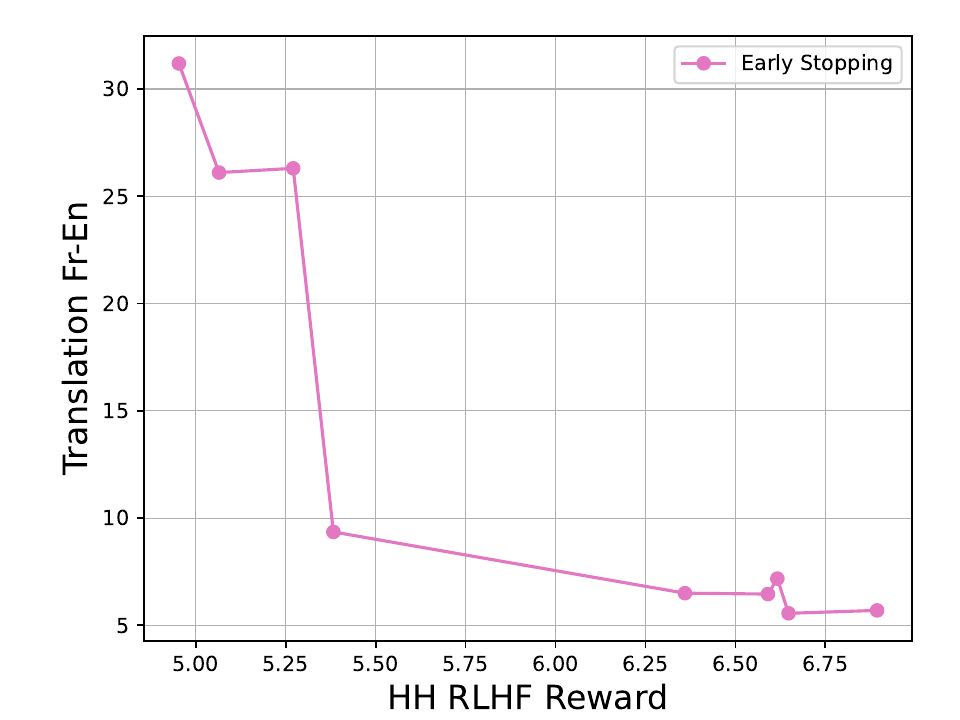}
    \caption{The alignment-forgetting trade-off during training}
    \label{fig:reward_tax_early_stopping}
\end{figure}
\subsection{More Results of Averaging Different Parts}
\label{app:more_ave_diff_parts}
In this part, we include the full results (e.g., RSF, DPO, PPO) of averaging different parts.
\begin{figure}[H]
    \centering
    \includegraphics[width=0.3\linewidth]{fig/RAFT-Split-Type1/hhrf_reward___drop_,__squad2___f1_tradeoff.pdf}
    \includegraphics[width=0.3\linewidth]{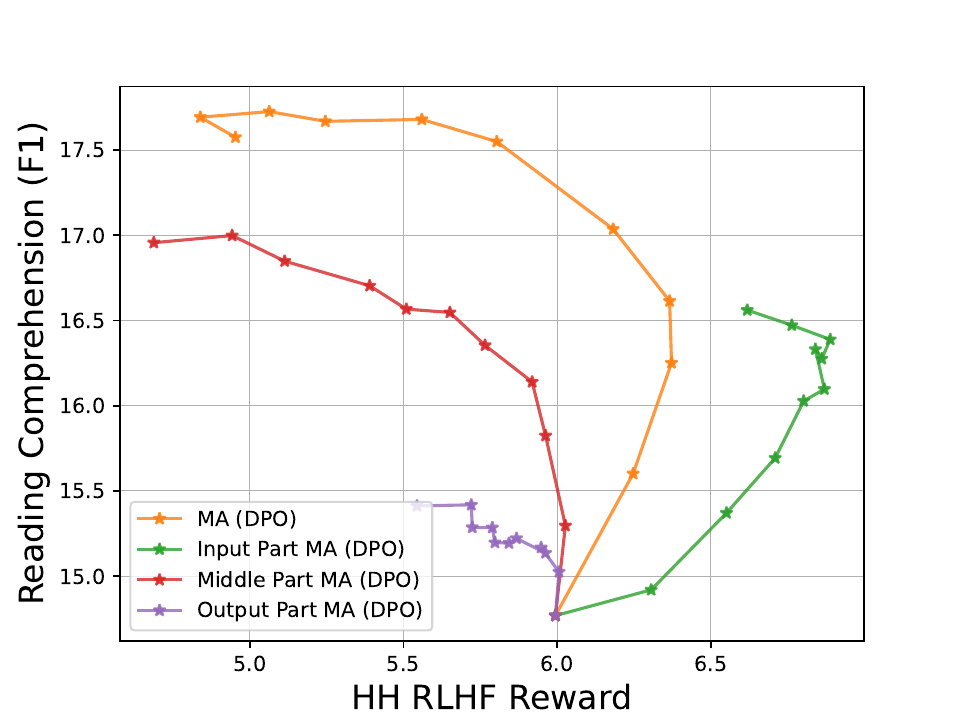}
    \includegraphics[width=0.3\linewidth]{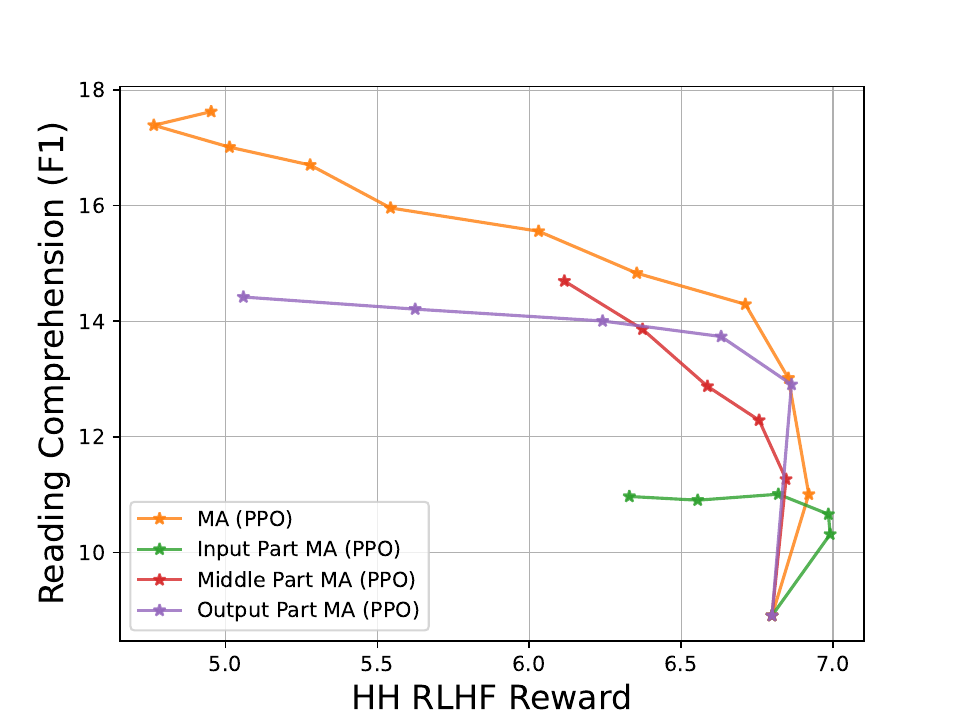}

    \caption{The performance of averaging different parts. (Left) RSF; (Middle) DPO; (Right) PPO}
    \label{fig:averaging_different_parts_full}
\end{figure}
\subsection{Comparison of RLHF Algorithms}
\begin{figure}[H]
    \centering
    \includegraphics[width=0.3\linewidth]{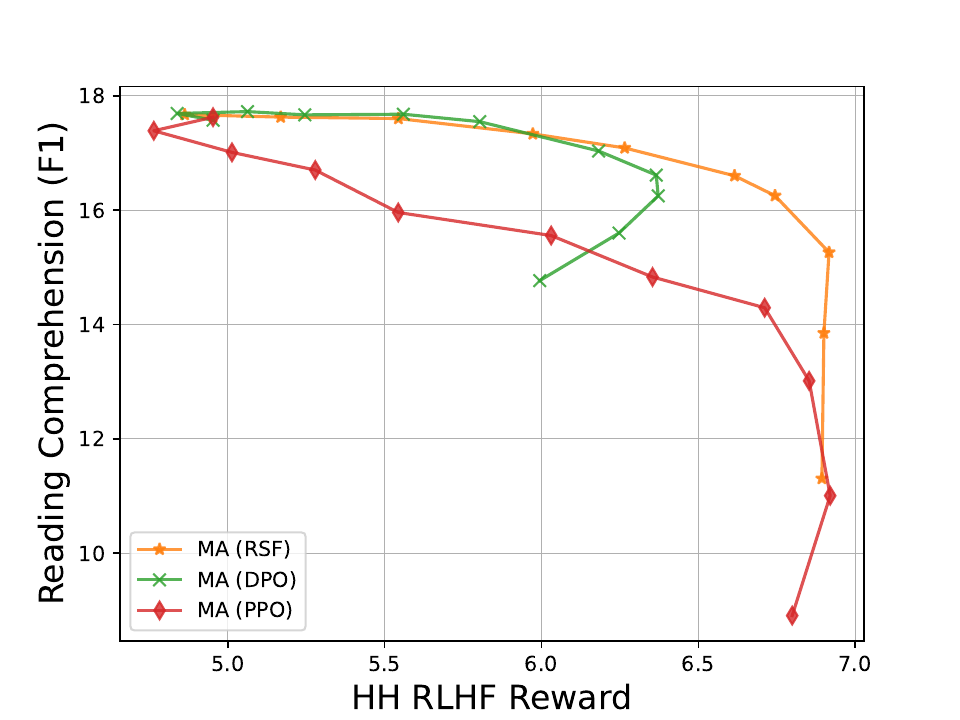}
    \includegraphics[width=0.3\linewidth]{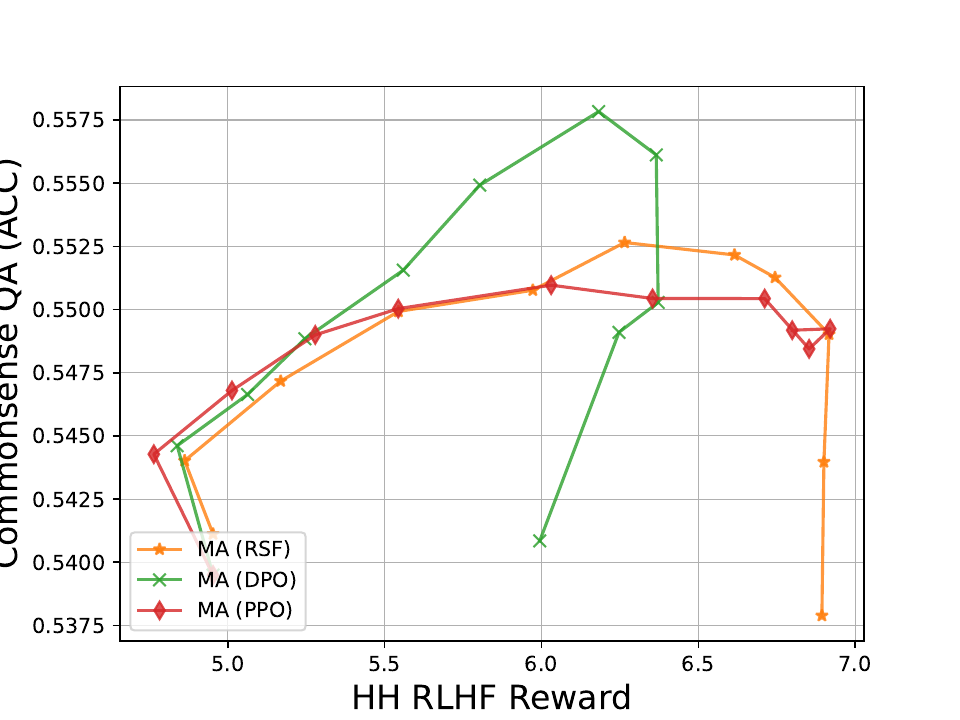}
    \includegraphics[width=0.3\linewidth]{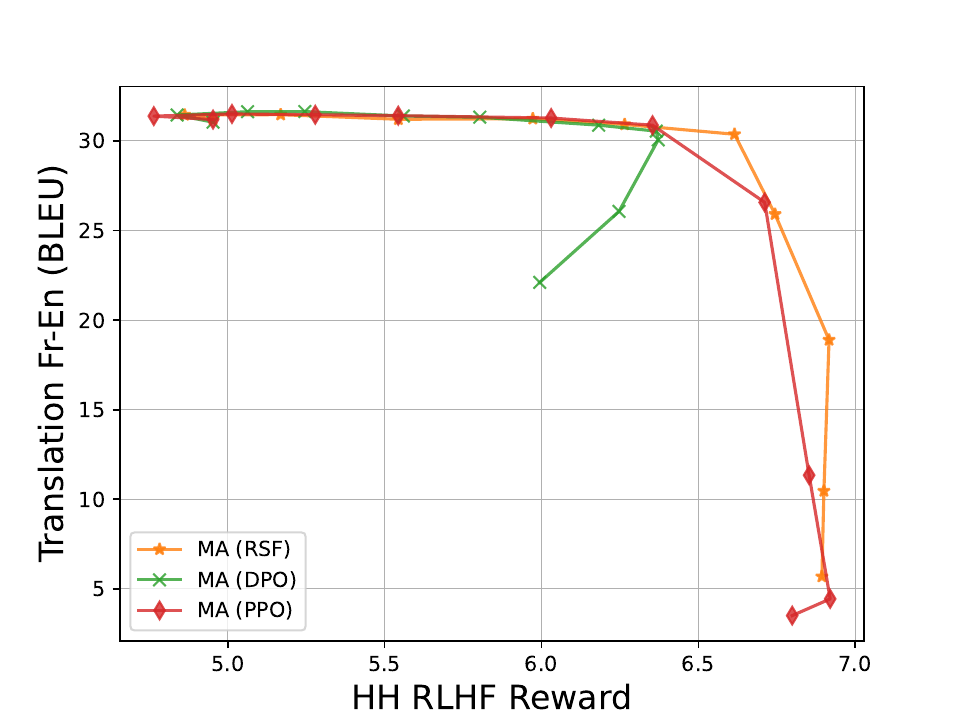}
    \caption{Comparison of RLHF algorithms in terms of alignment-forgetting trade-off.}
    \label{fig:comparision_of_RLHF_algorithm}
\end{figure}

We compare the alignment-forgetting trade-off of RSF, DPO and PPO in Figure~\ref{fig:comparision_of_RLHF_algorithm}. We observe that RSF is consistently better than DPO. However, we also note that this is not a fair comparison since DPO does not directly optimize for the reward.

\subsection{Results of AdaMerging \cite{yang2023adamerging}}
\label{app:ada_merging}
Previous studies \cite{yang2023adamerging} have also discussed the idea of dynamically assigning different weights to different layers when merging models, aiming to maximize performance on a specific task (e.g., $\cT_i$). These approaches assume access to the task-specific data $\cT_i$. However, considering the nature of alleviating alignment tax, which aims to mitigate forgetting across a extremely wide range of tasks ($\cT_{j_1}...\cT_{j_K}$), these methods fail to effectively optimize performance for multiple tasks simultaneously. Specifically, we want to preserve the abilities on a wide range of tasks and it is hard to get the data for all these tasks. Further more, some ability such as in-context learning does not have a clear corresponding training set. So it is less practical to find training set for AdaMerging. 

Here we demonstrate when we use AdaMerging to  optimizes for  task A and the training set does not cover task B, AdaMerging can not preserve the ability on task B.  Specifically, we provide AdaMerging with labeled data for Reading Comprehension (i.e., task A) and optimize the 26 layer-wise merging ratios as \cite{yang2023adamerging}. To have a clear comparison with vanilla model averaging, we try different mean averaging ratio for AdaMerging among 0.2, 0.4 and 0.6. We also show both the results on task A and B.
\begin{figure}
    \centering
    \includegraphics[width=0.3\linewidth]{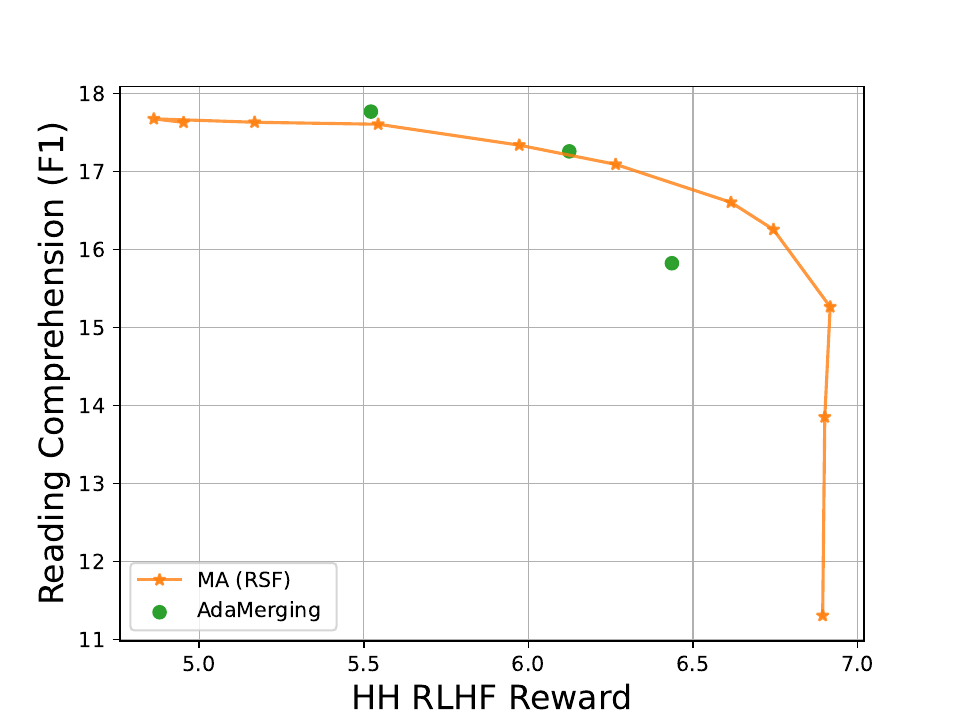}
    \includegraphics[width=0.3\linewidth]{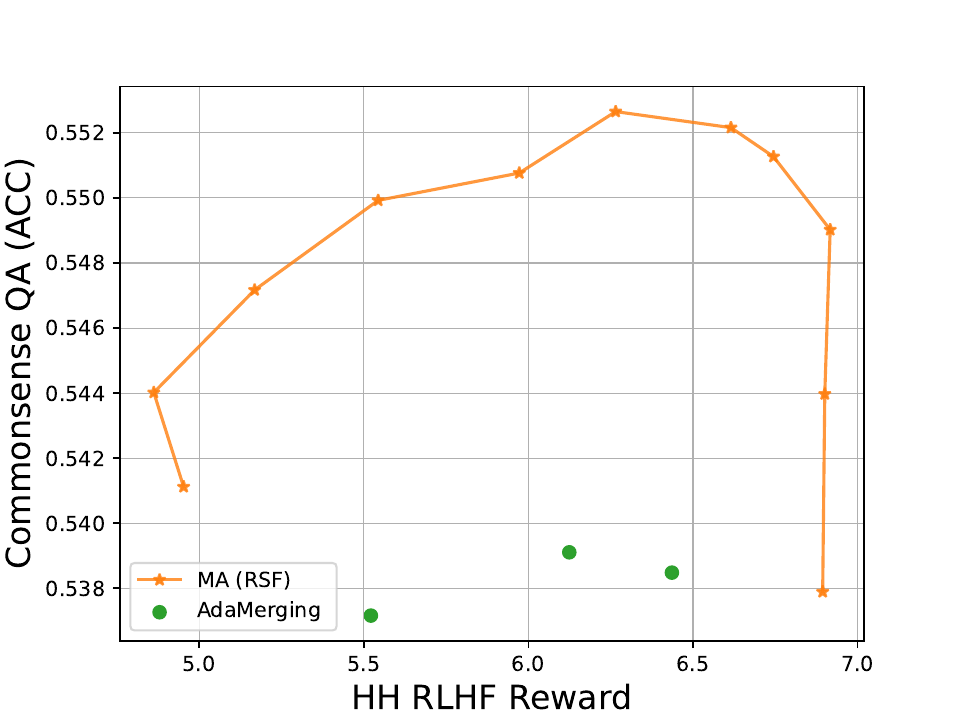}
    \caption{Results of AdaMerging. We optimize AdaMerging on Reading Comprehension and found it can hardly do well on Common Sense.}
    \label{fig:Ada_merging}
\end{figure}

\begin{figure} [H]
    \centering
    \includegraphics[width=0.3\linewidth]{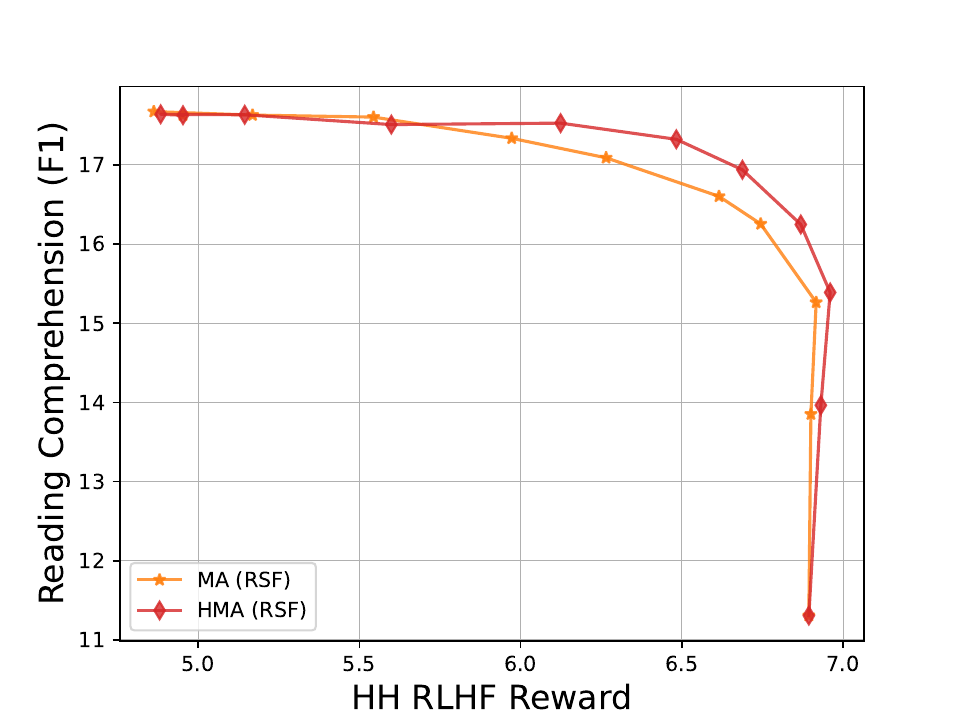}
    \includegraphics[width=0.3\linewidth]{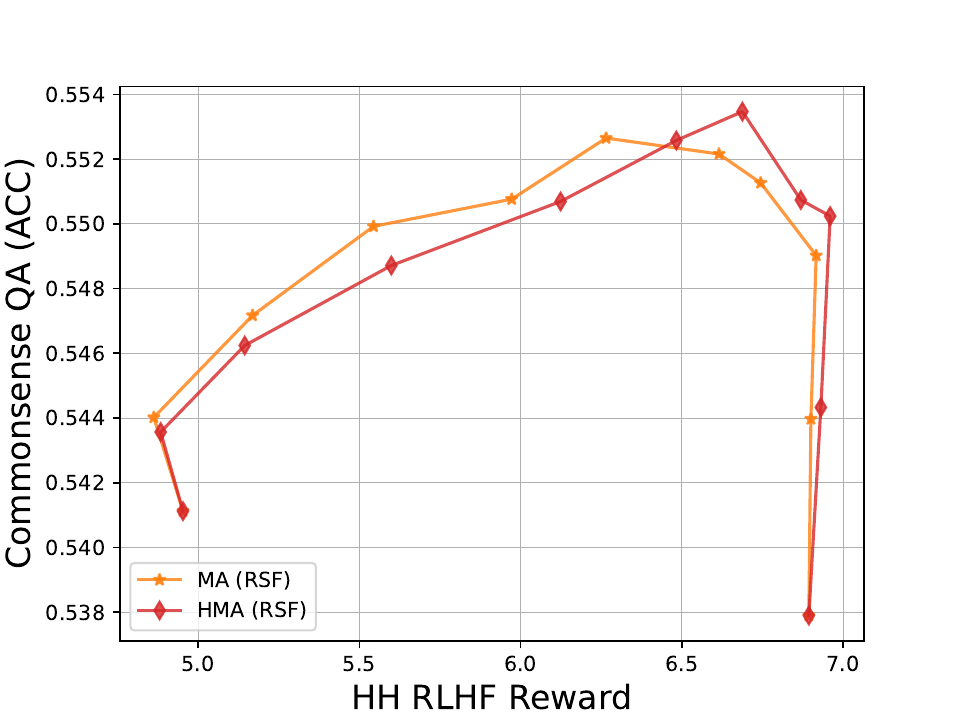}
    \includegraphics[width=0.3\linewidth]{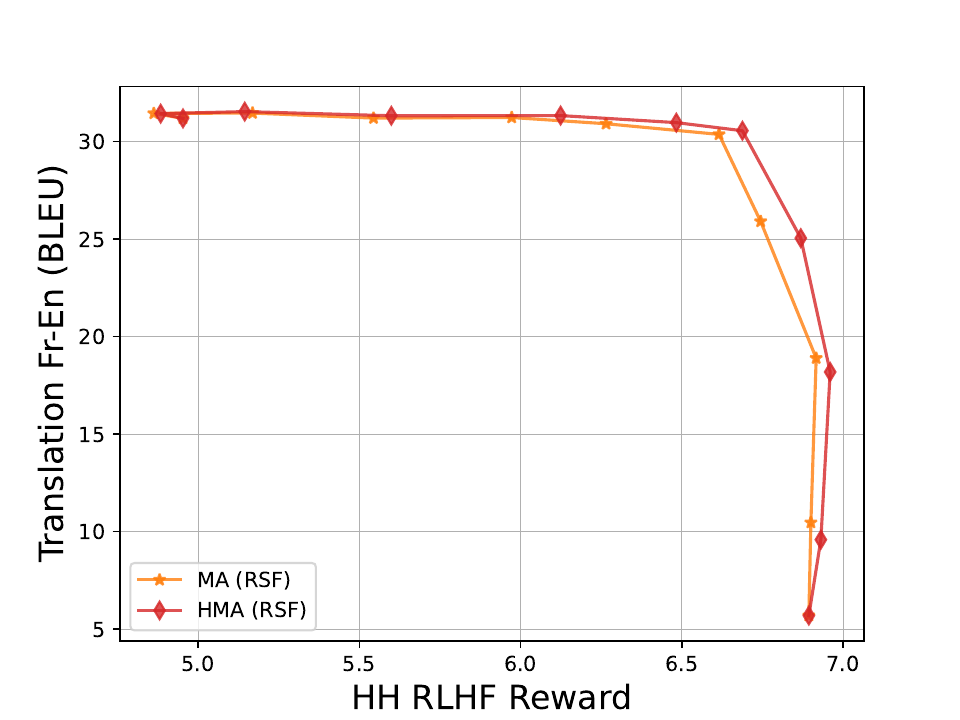}
    \includegraphics[width=0.3\linewidth]{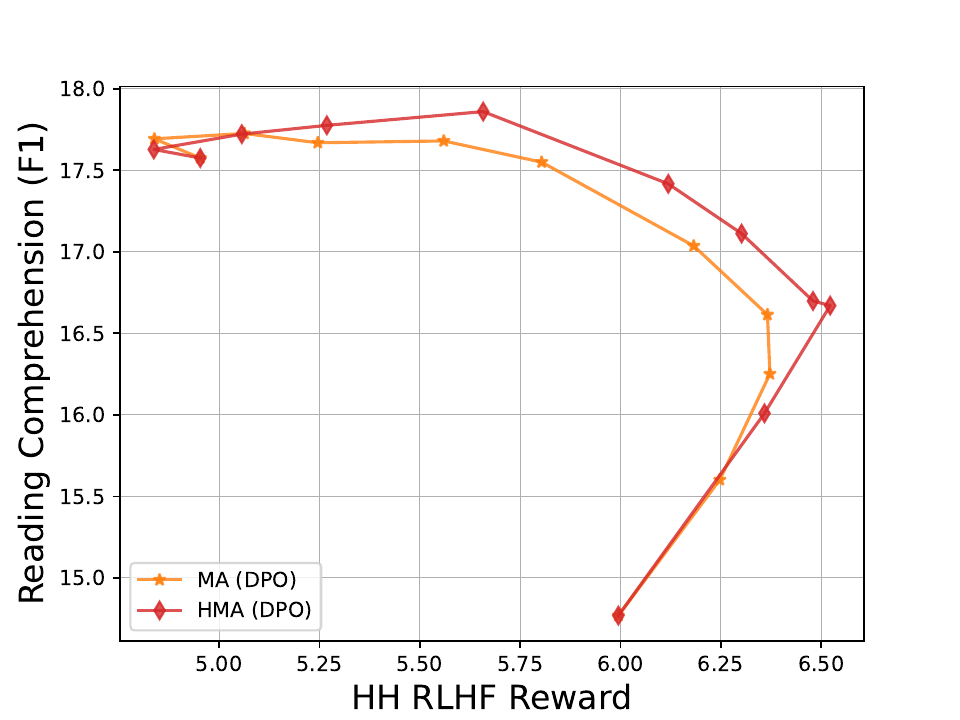}
    \includegraphics[width=0.3\linewidth]{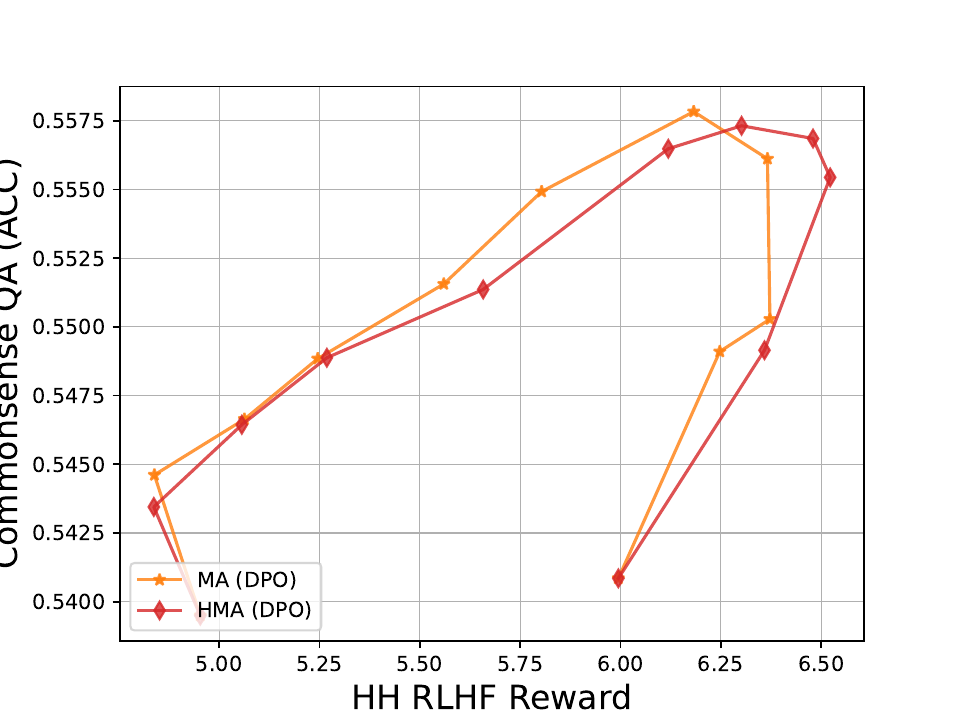}
    \includegraphics[width=0.3\linewidth]{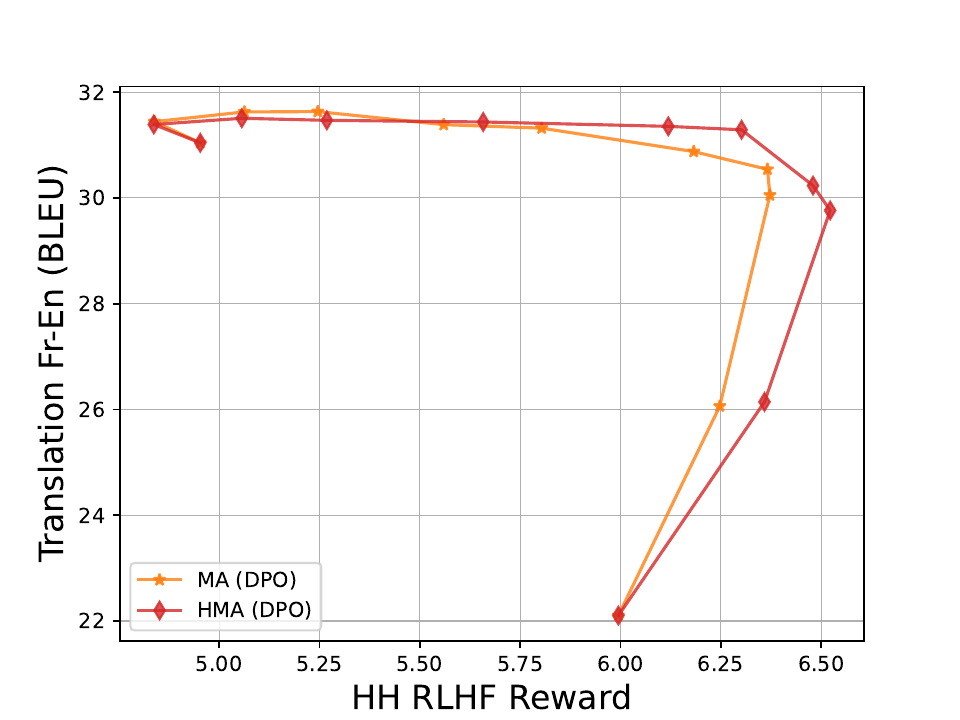}
    \caption{Detailed results of Heterogeneous model averaging on various benchmarks and RLHF methods.}
    \label{fig:detailed_results_ama}
\end{figure}

In contrast, our HMA only require the RLHF data and does not need any data from the tasks which we want to preserve ability. Figure~\ref{fig:detailed_results_ama}  shows that HMA can alleviate the alignment tax evaluated on a wide range of tasks.

\subsection{Detailed Results of Heterogeneous Model Averaging}
\label{app:detailed_HMA}

We provide the detailed results of Heterogeneous model averaging on various benchmarks, e.g., Reading Comprehension, Commonsense QA and translation, and different RLHF methods, e.g., RSF, PPO, and DPO.

\section{Theoretical Settings, Proofs and Discussions}
\subsection{Re-statement of Formal Settings}
\textbf{Notation}. Consider that the full class space $\cM$ contains $M$ classless, i.e. $\by \in \{\be_1, \be_2, ..., \be_M\}$, where $\be_i$ denotes the $M$-dimensional unit vector with $i$th element equaling 1, e.g., $\be_2 = [0, 1, 0, ..., 0]^\top$. $\boldsymbol{a}(k)$ means the $k$th element of vector $\boldsymbol{a}$, $\boldsymbol{A}(k)$ means the $k$th column of matrix $\boldsymbol{A}$.
We use  $\boldsymbol{I}_M$ to represent a $M \times M$ identity matrix, e.g., $\bI_M = [\be_1, \be_2, ..., \be_M]$. We omit the subscript of $\bI$ when no confusion arises.

Following \citep{lin2023spurious}, suppose we have $N$ weak features $\{\bx_i\}_{i=1}^{N}$ where $\bx_{i}  \in \bbR^d$ and the whole feature $\bx \in \bbR^{  d \times N  }$ is the concatenation of them, 
i.e., $\bx = \mbox{Concat} \Big(\{\bx_{ i}\}_{i=1}^{N} \Big) = [\bx_{1}, \dots, \bx_{N}].$ Consider that each model $f$ is composed of a featurizer $\Phi \in \{0, 1\}^{N}$ and a classifier $\bw \in \bbR^{d \times K}$. 
$\Phi$ first selects feature by $\bx \Phi$. For example, suppose $\bx = [\bx_1, \bx_2, \bx_3]$ and $\Phi = [1, 1, 0]^\top$, then $\bx \Phi = \bx_1 + \bx_2$.
Then the classifier $\bw \in \bbR^{d \times K}$ is fit based on the features selected by $\Phi$ as $\bw =  \argmin_{\bv }  \bbE[\ell(\bv^\top (\bx \Phi) , \by)] + \| v\|_2^2$,
where $\ell$ is the cross-entropy loss function.

We simplified \citep{lin2023spurious}'s Definition 1 and only consider weak features as following:

\begin{definition}[Data Generation Process]
    \label{defi:data_generation}
    The whole data generation process is as follows:
    \begin{align}
         & \boldsymbol{y} \sim \text{Unif} \left\{ \be_1, \be_2, ... \be_M \right\}, \bx = \mbox{Concat} \Big(\{\bx_{ i}\}_{i=1}^{M}\Big), \nonumber                          \\
         & \mathbb{P}_{\theta} (\boldsymbol{x}_{i} \mid \boldsymbol{y}) = \mathcal{N} \left( {\boldsymbol{\mu}}_{i}\bQ_{ i}\by, \sigma^2 \boldsymbol{I}_d \right), \forall i.
    \end{align}
    where $\bQ_{i} \in \{0, 1\}^{M \times M}$. the $m$th column of $\bQ$, i.e., $\bQ_{j}(m)$, is as follows for $m=1,2,\cdots,M$:
    $$\bQ_{j}(m) = \begin{cases}
            \be_m,  \mbox{ with probability } 1-p \\
            \mbox{Unif} \{\be_1, \cdots, \be_M\}, \mbox{ with probability } p.
        \end{cases}$$
\end{definition}

\begin{definition}
    [Model Averaging, Definition 4 of \citep{lin2023spurious}]
    \label{defi:model_ensemble}
    Given the two individual models $(\bar w, \bar \Phi)$ and  $(\tilde w, \tilde \Phi)$ ,   the prediction of the  model averaging is $f_{\\avg}(\bx) =  \frac{1}{4}(\bar \bw + \tilde \bw)^\top \left(\bx(\bar \Phi + \tilde \Phi)\right)$

    We impose the following mild assumptions as \citep{lin2023spurious}.
    \begin{assumption}[Small Noise]
        \label{ass:small_noise}
        Denote  $N_s$ as the the maximum number of invariant features and  spurious features that a model can learn, respectively.
        We need the overall noise to be small to satisfy $\boldsymbol{F}^{K}(\frac{1}{\sigma(N_s)}) \ge 1 - \epsilon,$
        in which $\boldsymbol{F}$ is the cumulative distribution function of standard Gaussian random variable, and $K$ refers to the class number.
    \end{assumption}

    \begin{assumption}[Orthogonal features \citep{lin2023spurious, allen2020towards}]
        \label{ass:ortho_feature}
        (1) $\| \boldsymbol{\mu}_{i}(k)\|_2 = 1$ for $i=1,\cdots, n$, (2) ${\mu}_{i}(k) \perp
            {\mu}_{i'}(k')$ for any $(i, k) \neq (i', k')$, $k, k'= 1, \cdots, K, $  $i, i' \in 1,\cdots, n$.
    \end{assumption}
\end{definition}
\subsection{Proof of Proposition~\ref{prop:intuitive_analysis}}
\paragraph{Estimating $\xi^{(1)}$ corresponding to Case (1).}
The estimation of $\xi^{(1)}$ is a direct application of Proposition 7 of \citep{lin2023spurious}. Specifically, according to Proposition 7 of \citep{lin2023spurious}, we have
\begin{align}
    \cA_a(f_a) = \cA_b(f_b) = F_b((1-p)\sqrt{n}),
    \cA_a(f_{\avg}) = \cA_b(f_{\avg}) = F_b((1-p)\frac{\sqrt{2} n}{ \sqrt{n + n_o}})
\end{align}

\paragraph{Estimating $\xi^{(2)}$ corresponding to Case (2).}

Without loss of generality, we assume the $\cY_a$ is $\{1, ..., K\}$ and $\cY_b$ is $\{K+1, ..., 2K\}$. Denote the feature learnt by $(\bw_a, \Phi_a)$ and $(\bw_b, \Phi_b)$ as $\bx_1, ..., \bx_n$ and  $\bx_{n - n_{o} + 1}, ...,\bx_n, ... \bx_{2n - n_{o}}$.
Since $\cA_a(f_{\avg}), \cA_b(f_{\avg}) \geq 0$, we trivially have $\xi^{(1)} \geq -F_p((1-p))\sqrt{n}$ by combing Proposition 7 of \citep{lin2023spurious}.

According to the Lemma 5 of \citep{lin2023spurious}, we have
\begin{align*}
    \bar \bw_a(k) = \sum_{i=1}^n {\mu}_{i}(k), \forall k = 1, \cdots, K,  \quad \bar \bw_b(k') = \sum_{i=n - n_{o} + 1}^ {2n - n_o} {\mu}_{i}(k'), \forall k' = K+1, \cdots, 2K,.
\end{align*}
We first estimate the accuracy of $f_{\avg}$ on task ($a$), i.e., $\cA_a(f_{\avg})$, for a sample from class $k \in 1, \cdots, K$ and $k' \neq k, k' \in 1, \cdots, K $. Then by $|\cY_a \cap \cY_b| = 0$ and Assumption~\ref{ass:ortho_feature}, we have
\begin{align*}
    (\bw_a(k) + \bw_b(k))^\top \bx (\bar \Phi_a + \bar \Phi_b)  |_{y=\be_k} = \bw_a(k)^\top \bx \bar \Phi_a +  \bw_b(k) \bx \bar \Phi_b |_{y=\be_k} = \bw_a(k)^\top \bx \bar \Phi_a|_{y=\be_k} \\
    (\bw_a(k') + \bw_b(k'))^\top \bx (\bar \Phi_a + \bar \Phi_b)  |_{y=\be_k} = \bw_a(k')^\top \bx \bar \Phi_a +  \bw_b(k') \bx \bar \Phi_b |_{y=\be_k} = \bw_a(k')^\top \bx \bar \Phi_a |_{y=\be_k}
\end{align*}
The last equality is due to $w_b(k)=0$ and $w_b(k')=0$ for $k, k' \in 1,...,K$. Then it is straightforward to see that $\cA_a(f_{\avg}) = \cA_a(f_a)$. We similarly have $\cA_b(f_{\avg}) = \cA_b(f_b)$. Then we have $\xi^{(2)} = 0$.

We finish the proof by collecting the results.

\subsection{Discussion on the Effect of Task Similarity on Model Averaging}
\label{app:theory_discuss}
We illustrate why model averaging would not lead to much improvement if two tasks are dissimilar, i.e., $|\cY_a \cap \cY_b| = 0$. Without loss of generality, we assume the $\cY_a$ is $\{1, ..., K\}$ and $\cY_b$ is $\{K+1, ..., 2K\}$.
Since $\bw$ is the minimum norm solution based on $\Phi$, we know that $\bw_b(k) = 0$ for $k=1, ..., K$. From the previous proof, we know that
\begin{align*}
    (\bw_a(k) + \bw_b(k))^\top \bx (\bar \Phi_a + \bar \Phi_b)  |_{y=\be_k} = \bw_a(k)^\top \bx \bar \Phi_a +  \bw_b(k) \bx \bar \Phi_b |_{y=\be_k}
\end{align*}
Since $\bw_b(k) = 0$, the above equation equals $\bw_a(k)^\top \bx \bar \Phi_a$, which is simply the performance of $f_a$. Intuitively, $\bw_b(k) \bx \bar \Phi_b$ maps the feature $\bx \bar \Phi_b$ into the space spanned by $\bw_b$. However, since $\bw_b$ is all zero in the dimension $1, ..., K$, so $\bw_b(k) \bx \bar \Phi_b$ has no impact on the prediction of task $a$ (i.e., among class $1, ..., K$).

\subsection{Close Form of $F_p(x)$}\label{app:fp}
Here we provide the explicit expression of function $F_p(x)$ in $K$ class situation, which is monotonically increasing with $x$.

We denote a $K-1$-dim random variable $\boldsymbol{\eta} \sim \mathcal{N}(\bx, \boldsymbol{M})$, in which 
\begin{align*}
& \boldsymbol{M}_{i,i} = \frac{p(K + 2 - pK)}{K}, \boldsymbol{M}_{i,j} = \frac{p(K + 1 - pK)}{K},
\end{align*}
then $F_p(x)$ is defined as
\begin{equation*}
    F_p(x) = \mathbb{P} (\boldsymbol{\eta}_1 > 0, \dots, \boldsymbol{\eta}_{K-1} > 0).
\end{equation*}

\section{Hyper-Parameters}

\begin{table}[htb]
    \centering
    \caption{Hyper-parameters for RLHF experiments with Open-LLaMA-3B. $\Delta$ means that the parameter will be specified in each individual experiment. For LoRA training, the omitted hyper-parameters are set as the full training. } \vspace{0.2in}
    \label{tab:hyper_exp_aux}
    \begin{sc}
        \begin{tabular}{c|c|c}
            \toprule
            Models and Methods  & Hyper-parameter            & Value             \\
            \midrule
                                & Temperature                & $1.0$             \\
                                & Data collection batch size & 512               \\
            PPO Training        & Learning rate              & $1\times 10^{-6}$ \\
                                & Update Epoch               & $2$               \\
                                & Update batch size          & $32$              \\
                                & KL coefficient             & $\Delta$          \\
                                & Reward baseline            & $5.5625$          \\
            \midrule
                                & Learning rate              & $1\times 10^{-5}$ \\
                                & Update Epoch               & $4$               \\
                                & Update batch size          & $32$              \\
            PPO LoRA Training   & KL coefficient             & $\Delta$          \\
                                & Reward baseline            & $5.5625$          \\
                                & LoRA rank                  & $16$              \\
                                & LoRA $\alpha$              & $32$              \\
                                & LoRA Dropout               & $0.05$            \\
            \midrule
                                & Temperature                & $1.0$             \\
            RSF Training        & Batch size                 & 2048              \\
                                & Learning rate              & $1\times 10^{-5}$ \\
                                & Epoch                      & $2$               \\
                                & Update batch size          & $32$              \\
            \midrule
                                & Learning rate              & $1\times 10^{-5}$ \\
                                & Epoch                      & $2$               \\
            {RSF LoRA Training} & Update batch size          & $32$              \\
                                & LoRA rank                  & 16-512            \\
                                & LoRA $\alpha$              &   32                \\
            \midrule
                                & Learning rate              & $1\times 10^{-6}$ \\
            DPO                 & Batch size                 & $32$              \\
                                & KL coefficient             & $0.1$             \\
            \bottomrule
        \end{tabular}
    \end{sc}
\end{table}

\begin{table}[htb]
    \centering
    \caption{Hyper-parameters for auxiliary experiments. }\vspace{0.2in}
    \label{tab:hyper_exp_aux1}
    \begin{sc}
        \begin{tabular}{c|c|c}
            \toprule
            Models and Methods & Hyper-parameter       & Value                          \\
            \midrule
                               & Learning rate         & $1\times 10^{-5}$              \\
                               & Scheduler             & Cosine decay with 0.03 warm-up \\
            ShareGPT SFT       & Epoch                 & 1                              \\
                               & Batch size            & 128                            \\
                               & Block size            & 2048                           \\
            \midrule
                               & Learning rate         & $1\times 10^{-5}$              \\
                               & Scheduler             & Cosine decay with 0.03 warm-up \\
            HH-RLHF SFT        & Epoch                 & 1                              \\
                               & Batch size            & 12                             \\
                               & Block size            & 2048                           \\
            \midrule
                               & Learning rate         & $2 \times 10^{-5}$             \\
            RM SFT             & Scheduler             & Cosine decay with 0.03 warm-up \\
                               & Epoch                 & 2                              \\
                               & Batch size            & 12                             \\
            \midrule
                               & Learning rate         & $5 \times 10^{-6}$             \\
            RM Training        & Scheduler             & Cosine decay with 0.03 warm-up \\
                               & Epoch                 & 1                              \\
                               & Batch size            & 16                             \\
            \midrule
                               & Temperature $\lambda$ & $1.0$                          \\
            Test Settings      & Max new token         & 196                            \\
                               & Do sample             & True                           \\
            \bottomrule
        \end{tabular}
    \end{sc}
\end{table}

\begin{table}[htb]
    \centering
    \caption{Hyper-parameters for HMA experiments. }\vspace{0.2in}
    \label{tab:hyper_exp_ama}
    \begin{sc}
        \begin{tabular}{c|c|c}
            \toprule
            Models and Methods & Hyper-parameter & Value                          \\
            \midrule
                               & Learning rate   & $2\times 10^{-5}$              \\
                               & Scheduler       & Cosine decay with 0.03 warm-up \\
            RSF HMA            & Epoch           & 1                              \\
                               & Batch size      & 1                              \\
                               & Block size      & 512                            \\
            \midrule
                               & Learning rate   & $4\times 10^{-5}$              \\
                               & Scheduler       & Cosine decay with 0.03 warm-up \\
            PPO HMA            & Epoch           & 1                              \\
                               & Batch size      & 1                              \\
                               & Block size      & 512                            \\
            \midrule
                               & Learning rate   & $4 \times 10^{-5}$             \\
            DPO HMA            & Scheduler       & Cosine decay with 0.03 warm-up \\
                               & Epoch           & 1                              \\
                               & Batch size      & 1                              \\
                               & Block size      & 512                            \\
            \bottomrule
        \end{tabular}
    \end{sc}
\end{table}

\end{document}